\documentclass[letterpaper]{article} 
\usepackage{aaai25}  
\usepackage{times}  
\usepackage{helvet}  
\usepackage{courier}  
\usepackage[hyphens]{url}  
\usepackage{graphicx} 
\usepackage{natbib}  
\usepackage{caption} 
\usepackage{algorithm}
\usepackage{algorithmic}

\usepackage{amsmath} 
\usepackage{color}
\usepackage{amssymb} 
\usepackage{ragged2e}
\usepackage{multirow}

\usepackage[dvipsnames]{xcolor}

\usepackage{tablefootnote}

\usepackage[most]{tcolorbox}
\newtcolorbox{bubble}[1][]{%
  enhanced,
  sharp corners,
  boxrule=0pt,
  left=4pt, right=4pt, top=4pt, bottom=4pt,
  #1
}
\usepackage{newfloat}
\usepackage{listings}
\DeclareCaptionStyle{ruled}{labelfont=normalfont,labelsep=colon,strut=off} 
\floatstyle{ruled}
\newfloat{listing}{tb}{lst}{}
\floatname{listing}{Listing}
\title{Fine-Grained Interpretation of Political Opinions in Large Language Models}
\author {
    Jingyu Hu\textsuperscript{\rm 1},
    Mengyue Yang\textsuperscript{\rm 1},
    Mengnan Du\textsuperscript{\rm 2},
    Weiru Liu\textsuperscript{\rm 1}
}
\affiliations {
    \textsuperscript{\rm 1}University of Bristol\\
    \textsuperscript{\rm 2}New Jersey Institute of Technology\\
    ym21669@bristol.ac.uk, mengyue.yang@bristol.ac.uk, mengnan.du@njit.edu, weiru.liu@bristol.ac.uk
}

\usepackage{bibentry}

\begin{document}

\maketitle

\begin{abstract}
Studies of LLMs’ political opinions mainly rely on evaluations of their open-ended responses.
Recent work indicates that there is a misalignment between LLMs’ responses and their internal intentions.
This motivates us to probe LLMs' internal mechanisms and help uncover their internal political states.
Additionally, we found that the analysis of LLMs' political opinions often relies on single-axis concepts, which can lead to concept confounds.
In this work, we extend the single-axis to multi-dimensions and apply interpretable representation engineering techniques for more transparent LLM political concept learning. 
Specifically, we designed a four-dimensional political learning framework and constructed a corresponding dataset for fine-grained political concept vector learning. These vectors can be used to detect and intervene in LLM internals. 
Experiments are conducted on eight open-source LLMs with three representation engineering techniques. Results show these vectors can disentangle political concept confounds. Detection tasks validate the semantic meaning of the vectors and show good generalization and robustness in OOD settings. Intervention Experiments show these vectors can intervene in LLMs to generate responses with different political leanings.
\end{abstract}

%

\section{Introduction}
Despite the success of large language models (LLMs) in many fields and tasks, there is growing public concern about the ethical implications of LLMs.
Many studies show that LLMs can replicate and even amplify societal biases \cite{bender2021dangers,wan-etal-2023-kelly}. 
Such social bias can appear as racial or gender stereotypes, or as a preference for certain values in responses. 
This issue becomes especially tricky and subtle in the political domain since LLMs favouring specific values can unconsciously influence the decisions of the policy-makers.

Some studies have revealed that LLMs exhibit systematic political biases and have attempted to evaluate and correct them \cite{rozado2024political, motoki2024more}.
\citet{piao2025emergence} designed LLMs-based multi-agent systems to simulate its potential social impacts, it was discovered that LLMs can exhibit opinion polarization phenomena similar to the human society. These assessments are generally prompt-based, directly asking LLMs about their left- or right-leaning position on some selected topics.
Recent research has begun exploring possibilities of assessing political bias via tuning LLMs' parameters. 
For example, PoliTune \cite{agiza2024politune} used left-leaning and right-leaning datasets for Direct Preference Optimisation (DPO) and ultimately produced two fine-tuned LLMs with opposing political preferences. This work indicates that the internal parameters of LLMs are highly sensitive to data selection with different viewpoints, and can be manipulated through data to favour a particular stance.

Nevertheless, these strategies still treat LLMs as black boxes, assessing their explicit political bias via generated outputs while neglecting the role of internal intentions.
\citet{marks2025auditing} show that LLMs can appear to achieve the intended objectives while their hidden intentions remain misaligned.
If such misalignment contains implicit hidden biases, those biases can subtly influence users' opinions: studies have found that approximately one-fifth of users can shift their values after interacting with LLMs \cite{ju2025benchmarking,potter2024hidden}.
This motivates us to explore LLMs’ internal mechanisms to understand how different political leanings emerge and change within LLMs.

Recent work on LLMs’ interpretability indicates the possibility of understanding and steering LLM behaviours by learning their internal feature representations \cite{turner2023activation,panickssery2023steering}. The core idea of representation engineering is to discover the hidden states of each layer of LLMs, train vectors representing concepts occurring in these representations, and use these vectors to detect LLMs’ internal state and steer LLM outputs to align with given concepts by intervention on specific representations. 
For example, one can train left-leaning or right-leaning vectors to steer an LLM’s outputs left or right. 
Although representation learning has been discussed in many fields, it is still largely unexplored in the politics domain. This paper aims to introduce a systematic exploration of LLMs’ internal political concepts via representation engineering techniques.

Meanwhile, there is a unique challenge in LLMs' political discussions different from other domains: most current work operates on political datasets within a single left-right axis, however, in real political scenarios, the boundary between left and right can be subtle, which can lead to conceptual confounds between `left' and `right'. 
For example, the political compass\footnote{\url{https://www.politicalcompass.org/}} notes that France’s National Front, popularly described as ‘far right’, actually supports left-leaning economic policies. Some recent work has tried to include a second axis (e.g., the progressive-conservative axis) to supplement the left-right spectrum; however, the essential issue remains unaddressed: what exactly do the concepts of `left' and `right' mean?
Due to cultural differences, definitions of left and right are not as clear as physiological concepts like gender or age. The `left is right' phenomenon \cite{wojcik2021left} can occur, leading to unconscious concept confounds. 
As the model is pre-trained on real-world data, LLMs can also contain similar inconsistencies, where content described as right-leaning might actually contain left-leaning views, causing LLMs to misassociate some right-leaning concepts with left-leaning ones.

To disentangle these conceptual confounds and enhance transparency in LLMs' political learning, our work proposes a fine-grained political learning framework with mechanistic interpretability techniques.
Specifically, we first construct datasets representing left- and right-leaning statements across four dimensions: economic, diplomatic, civil and society. Then interpretability techniques are applied to extract politically disentangled representations from these datasets, which enable us to derive corresponding political concept vectors. 
The concept vector correlation analysis shows we can correctly identify the disentangled concept representations.
We also designed a detection task to predict the political leaning for a given statement using the concept vectors. The detection performance verifies that the inferred concept vectors have semantic meaning and perform well in both in-distribution and out-of-distribution settings.
Further intervention experiments show that these concept vectors can modify the LLMs' internal representations and ultimately steer the models to generate content aligned with the leaning of the political concept vector.

Overall, this work aims to design a fine-grained framework based on representation engineering to more transparently and meticulously explore the internal political states of LLMs. The contributions can be summarized in three folds:
\begin{itemize}
    \item \textbf{Fine-grained Political Framework} We extended the single-axis political analysis to four dimensions and constructed the corresponding fine-grained political dataset to mitigate political concept confounds.
    \item \textbf{Internal Political Concept Vector Learning} 
Based on the designed framework, we applied representation engineering techniques to learn disentangled political concept vectors from LLM internals. The experiments were conducted across eight open-source LLMs using three interpretable representation engineering methods (CAA, RepE, and Linear Probing).
    \item \textbf{Robust Performance on Detection and Intervention Tasks} We evaluated the effectiveness of these concept vectors on LLM internals detection and intervention tasks.
The results show that these vectors achieved strong detection performance in both in-distribution and out-of-distribution settings. Moreover, these vectors can successfully intervene in LLM outputs to generate statements with different political leanings.
\end{itemize}

\begin{figure}[t]
    \centering
    \includegraphics[width=0.8\linewidth]{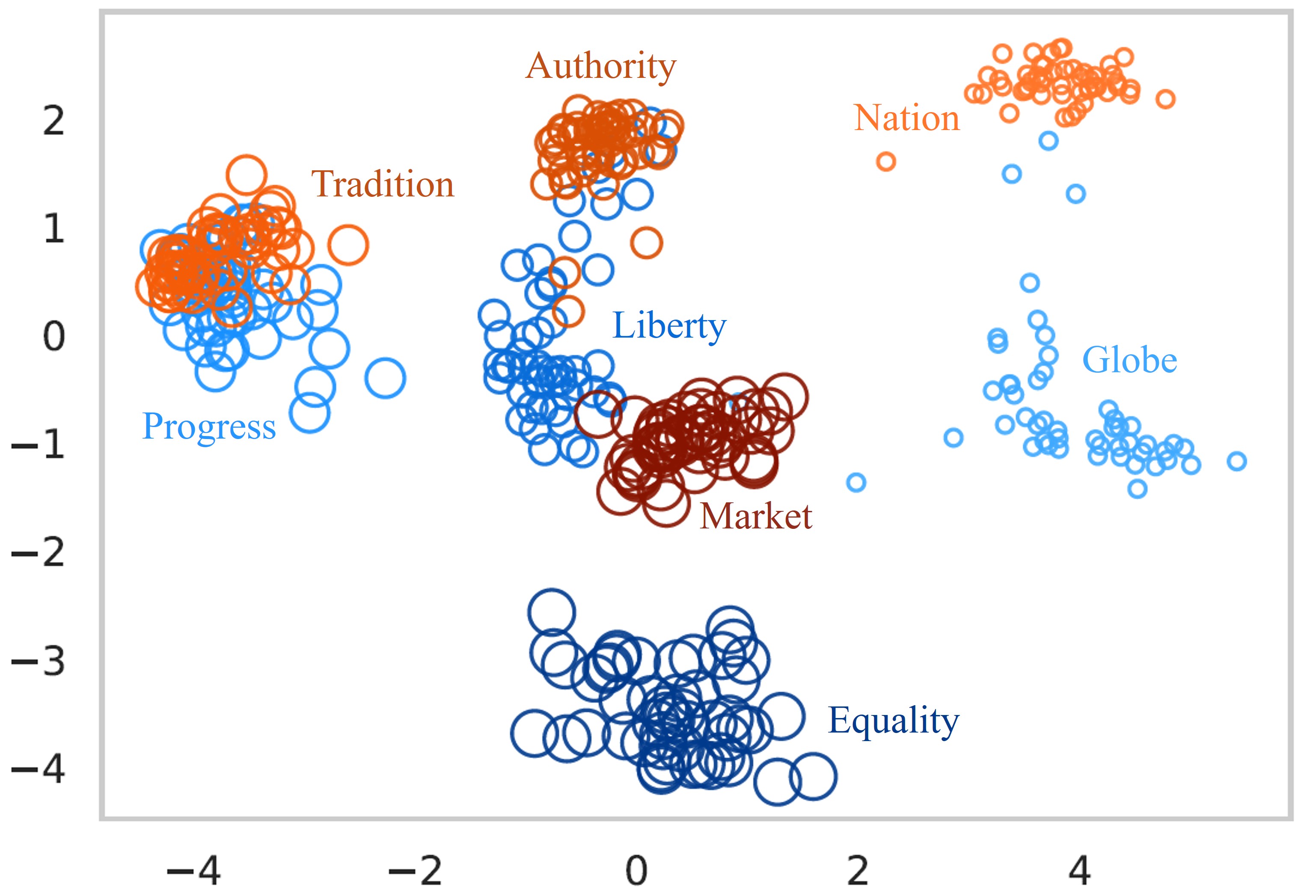}
    \caption{An Example of Left-Right Concept Confounds (PCA, Llama3-8b, Layer=20), where blue tones refer to left-leaning concepts and red tones represent right-leaning concepts. The right-leaning economic `Market' concept is more mixed with the left-leaning civil `Liberty' concept.}
\end{figure}

\section{Related Work}

\begin{table*}[t]
\centering
\caption{The Comparison of Recent Related Work on LLMs Political Opinions, where `--' refers to not applicable}
\label{tab:cmp_diff_poli}
\resizebox{0.97\textwidth}{!}{
\begin{tabular}{ccccccc}
\hline
           & \textbf{Methods}             & \textbf{Interpretability} & \textbf{Disentangled} & \textbf{Dimensions} & \textbf{Topics} & \textbf{Collected Data size}  \\ \hline
\cite{rottger2024political} & Prompt-based             & $\times$  & $\times$ & 2 & 6     &  --       \\
\cite{agiza2024politune}    & DPO Fine-tuning                & $\times$ & $\times$  & 2 & --    &  $\sim$5K  \\
\cite{kim2025linear}   & Probing             & \checkmark &  $\times$ & 1 & 9     &  $\sim$500 \\
(Ours)        & Probing, CAA, RepE     & \checkmark & \checkmark &  4 & 17    & $\sim$10K \\ \hline
\tiny
\end{tabular}}
\end{table*}

\begin{figure*}[t]
    \centering
    \includegraphics[width=0.95\linewidth]{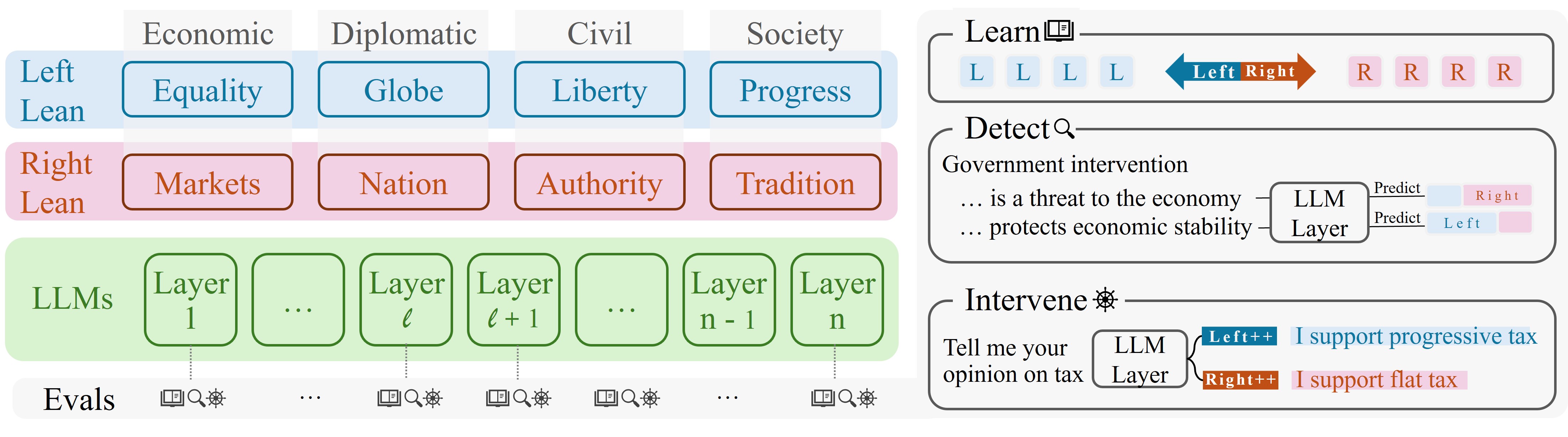}
    \caption{The Proposed Fine-grained Political Learning Within LLMs’ Internal States}
    \label{fig:enter-label}
\end{figure*}

\subsection{Interpretable Representation Engineering in LLMs}

Studies on LLMs' representation learning show that each layer's internal representations encode rich information like semantic content, concepts, factual knowledge, and attributes \cite{gurnee2023language,li2023inference,jin-etal-2025-exploring}. Also, the different sub-regions of the representation space appear to specialise in different tasks. This suggests the LLM's hidden states are learnable and interpretable, and has motivated growing interest in LLMs' mechanistic interpretability (MI). Activation representation engineering is one MI technique, with its core idea being to learn a set of concept vectors that can be used to detect and intervene in LLM internal states. Many concept vector learning algorithms including RepE \cite{zou2023representation}, CAA \cite{belinkov2022probing}, linear probing and other variations \cite{zhao2024beyond,chu2024causal,ousidhoum2021probing} have been proposed and widely discussed \cite{li2023inference,tak2025mechanistic,huang2025steering}. These methods essentially follow a similar process: given sets of positive and negative prompts that respectively present a desired information/function and their opposite as the dataset, the activation vector can be calculated by applying methods to learn the representation differences between prompt pairs within LLMs' each layer.

Detection and intervention are two types of tasks for evaluating the effectiveness of learned vectors. 
Reading includes locating emergent representations of the desired information in LLMs' internals.
The detecting quality is usually evaluated by performance on the designed classification task, with related discussions including detecting LLMs' hallucination \cite{ji2024llm}, deception \cite{goldowsky2025detecting}, and fact-checking ability \cite{marks2023geometry,he2024llm}.
Intervention is related to activation steering \cite{turner2023activation}, which injects learned vectors back into LLM internal representations during its forward pass at certain layer(s). Recent studies have shown this intervention can steer LLMs' internal states and guide LLMs to generate desired responses \cite{dathathri2019plug,subramani2022extracting}. 
Related discussions on steering include many aspects like psychology \cite{tak2025mechanistic}, safety \cite{ball2024understanding,xu2024uncovering}, interactive applications\cite{turner2023activation}, unified framework \cite{bhalla2024towards,wu2024pyvene,im2025unified}.
The intervention quality is generally evaluated through designed metrics based on user experience, LLM-as-a-Judge score, or variations in output token logits score \cite{pres2024towards,turner2023steering}.
To better learn, detect, and steer the political states of LLMs, this work aims to apply these techniques to the political domain—where interpretability remains largely unexplored.

\subsection{Social Bias Impact and Mitigations in LLMs}

LLMs can replicate and even amplify social biases \cite{bender2021dangers,hu-etal-2024-strategic,tan2025unmasking}. 
In the political area, \cite{rozado2024political} finds that GPT models lean towards the political left. \cite{motoki2024more} further reveals that LLMs show that LLMs show systematic bias in favor of certain political parties.

Based on these findings, some work has begun to explore the social impact of such biases.
\cite{piao2025emergence} used large-scale multi-agent simulations and demonstrated that, when thousands of LLM agents interact and exchange political opinions over multiple rounds, they become more inclined to communicate with agents holding the same opinions, exhibiting similar polarization and echo-chamber phenomena to those found in human society.
\cite{potter2024hidden} examined LLMs’ political leanings and their potential influence on voters through multiple experiments framed around a U.S. presidential election.

Other work focuses on mitigating political bias.
PoliTune \cite{agiza2024politune} applies Direct Preference Optimisation (DPO) to fine-tune identical base models into `left-leaning' and `right-leaning' variants, showing LLMs' direction of political opinions are sensitive to the fine-tuning corpus.
Still, quantifying the political leanings embedded in LLMs' outputs remains difficult, as ideological differences are often subtle and highly context-dependent.
\cite{paschalides2025probing} introduces a voting-based ranking approach to explore the subtle ideological exposes and manipulation. 
These discussions often evaluate LLMs' bias in the explicit way, and little work has been done to prob LLMs’ internal political states. \cite{kim2025linear} is the only work that attempts to use probe LLMs’ political concepts, but like other studies, it collects data with coarse single-axis left-right leans to align with mainstream party positions, and overlooks potential concept confounds.
There is a lack of a fine-grained discussion on LLMs' internal political concepts' emergence and how we can intervene in them transparently. The comparison between our work with recent related work is illustrated in Table \ref{tab:cmp_diff_poli}.

\section{Methodology}

\begin{table*}[t]
\centering
\caption{An Overview of the Collected Dataset Component}
\label{table:ds_stas}
\resizebox{0.97\textwidth}{!}{
\begin{tabular}{cccccc}
\hline
\textbf{Notation} & \textbf{Dimensions}   & \textbf{Concepts} & \textbf{No. Topics}                                      & \textbf{Topics}     \\
\hline
$D^{\text{eco}} $ & Economic &Equality (L) / Market (R)   & 3          & Economy and jobs, taxes, banking and finance \\
$D^{\text{civil}} $ & Diplomatic & Globe (L)  / Nation (R)  & 4          & World, immigration, foreign policy, politics   \\
$D^{\text{dip}} $ & Civil  &Liberty (L)  / Authority (R) & 5 & Civil rights, voting rights, gun, abortion, free speech      \\
$D^{\text{soc}} $ & Society & Progress (L) / Tradition (R) & 5 & Technology, religion and faith, education, culture, LGBTQ \\
\hline
\tiny
\end{tabular}}
\end{table*}

\begin{figure*}[t]
    \centering
    \includegraphics[width=0.9\linewidth]{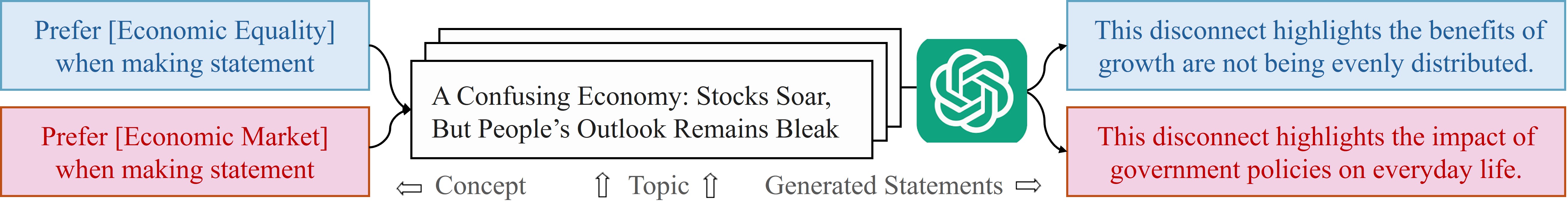}
    \caption{An Illustration of Constructing Left-leaning and Right-leaning Statements at Economic Dimension}
    \label{fig:example_generated}
\end{figure*}

\subsection{LLMs Architecture}
LLMs response process can be intuitively viewed as applying a pre‐trained neural network $f$ that maps the initial sentence (prompt) to a distribution over the collection of possible choices for the next token, then extends the sentence autoregressively to form the response.
We refer to \cite{duetting2024mechanism,chatzi2024counterfactual} for a formal description of this process

\textbf{Autoregressive Token Generation} Given a prompt $p_0$ which provides instructions for LLM generation, the tokenizer first encodes $p_0$ into a model-readable token sequence \(s_0\) according to the LLM token vocabulary $\mathcal{V}$. 
The LLM produces the next token for $s_0$ by sampling a new token \(\tau_1 \sim f(s_0)\) and constructing \(s_1 = s_0 \oplus \tau_1\) (where \(\oplus\) denotes concatenation). It then repeats the process by sampling \(\tau_k \sim f(s_k)\) and setting \(s_k = s_{k-1} \oplus \tau_k\) until a predefined stopping criterion is reached (e.g., the $k$-length limit, an end-of-sentence token). The tokenizer eventually decodes $s_k$ into human-readable text based on $\mathcal{V}$ as the LLMs response.

\textbf{Layer-wise Forward Pass}
The common architecture of $f$ is a decoder‐only transformer pre‐trained on massive natural language datasets. Specifically, $f$ consists of $n$ layers for representation inference. Each layer $\ell \in \{1,\dots,n\}$ shares the similar structure, including a multi‐head attention block $\mathrm{MHA}^\ell$ and a feed‐forward network block $\mathrm{FFN}^\ell$.
Given $h_{s}^{\ell}$ as the hidden representation of a token sequence $s$ at layer $\ell$, the hidden representation at the next layer $h_{s}^{\ell+1}$ is calculated as follows:

\begin{equation}
h_{s}^{\ell+1} = h_{s}^\ell + \mathrm{MHA}^\ell\bigl(h_{s}^\ell\bigr) + \mathrm{FFN}^\ell\!\bigl(h_{s}^\ell + \mathrm{MHA}^\ell(h_{s}^\ell)\bigr).
\end{equation}

The forward pass of $f(s)$ includes passing these embeddings layer by layer, then transforming and normalising the last layer $h_{s}^{n} $ back to the space with $\mathcal{V}$ size, from which we obtain tokens sampling distribution over the next token.

\subsection{LLMs Representation Engineering}

LLMs representation engineering usually includes learning, detecting and intervening in the hidden states within the LLMs. Our objective is to learn political opinion vectors via datasets with contrastive political statements, and use the learned vectors for detecting and intervening in LLM political opinions.

\subsubsection{Data Structure}
To better describe the structure of our constructed data, we first introduce three terms: dimensions set(\text{Dim}), concepts set($C$), and topics set($T$).
The dimensions set refers to our four fine-grained dimension set—Economic, Diplomatic, Civil, and Society—and can be denoted as $\mathrm{Dim}\mathord{=}\{\mathrm{eco},\mathrm{dip},\mathrm{civil},\mathrm{soc}\}.$
The discussion of concepts and topics takes place within each dimension $d \in \text{Dim}$. Specifically, concepts set is denoted as $C^{d}=\{C^{d}_{L},C^{d}_{R}\}$, which refers to the left-leaning and right-leaning definitions within $d$.
The topics set can be written as $T^d = \{\,T^d_\text{t}\}_{\text{t}=1}^m$, which refers to $m$ themes related to dimension $d$.
For example, in the economic dimension ($d \mathord{=} \mathrm{eco}$), concept $C^d$ can be economic equality ($C^{d=\mathrm{eco}}_L$) and market ($C^{d=\mathrm{eco}}_R$), and possible economic topics can be jobs ($T_1^{d=\mathrm{eco}}$), taxes ($T_2^{d=\mathrm{eco}}$), banking ($T_3^{d=\mathrm{eco}}$).

Our constructed dataset covers all dimensions and is denoted as ${\mathcal{D} = \{ D^{\mathrm{eco}} \cup D^{\mathrm{dip}} \cup D^{\mathrm{civil}} \cup D^{\mathrm{soc}} \} }$. 
Each constituent set $D^{d}\subseteq\mathcal{D}$ contains both left-leaning statements and right-leaning statements within dimension $d$. Specifically, these statements combine the corresponding left-leaning concept $C^{d}_{\text{L}}$ or right-leaning concept $C^{d}_{\text{R}}$ with the given topic $T^{d}_{\text{t}}$ to reflect different political learning statements on the given dimension and topic.

\begin{equation}
D^{d} = \{S^{d}_{\text{L,t}}, S^{d}_{\text{R,t}}\}^m_{\text{t}=1},
\end{equation}

\begin{equation}
S^{d}_{\text{L,t}} = C^{d}_{\text{L}} \odot T^{d}_{\text{t}}, \space
S^{d}_{\text{R,t}} = C^{d}_{\text{R}} \odot T^{d}_{\text{t}}.
\end{equation}
Here $\odot$ refers to the combination of the concept and topic. The implementation details are discussed in the experimental section.

\subsubsection{Political Concept Vector Learning}
Based on the representation linear hypothesis \cite{park2023linear,elhage2022toy}, which suggests that concepts are encoded linearly within model representations, many methods have been proposed to learn a concept vector $\vec{u}^{\,\ell}$ that encodes the underlying LLMs concept information. CAA \cite{belinkov2022probing}, RepE \cite{vogel2024repeng,zou2023representation} and Linear Probing \cite{ousidhoum2021probing} are three techniques used in this paper for political concept vector learning. 
To avoid concept confounds, we implement these methods to learn the concept vector $\vec{u}^{\ell}$ in different dimensions. Our insight is that LLM representations of data from different dimensions are disentangled (no overlap across dimensions). The disentangled concept vectors thus can be learned from the disentangled representations. The method details are as below.

Both CAA and RepE learn concept vectors through pairs of contrastive statements.
For each contrastive pair $(S_L, S_R)$ on topic $T^d_t$ of dimension $d$, we denote its corresponding left-leaning statement as $S_L$ and the right-leaning statement as $S_R$, where we assign $S_L := S^d_{L,t}$ and $S_R := S^d_{R,t}$.
Intuitively, the hidden representations difference between these two $\text{Diff}(h^{\ell}_{S_\text{L}},h^{\ell}_{S_\text{R}})$ 
captures how the LLM distinguishes the `left' and `right' differences at $\ell$-th layer.

CAA calculates the embeddings difference for each contrastive pair $(S_{\text{L}}, \,S_{\text{R}})$, and takes the weighted mean of these differences as concept vector $\vec{u}^{\ell}$, where

\begin{equation}
\label{eq:md-vector}
\vec{u}^{\,\ell}
\;=\; \sum_{S_L\in\mathcal{D}^d}\frac{h_{S_L}^{\,\ell}}{|\mathcal{D}^d|}
\;-\; \sum_{S_R\in\mathcal{D}^d}\frac{h_{S_R}^{\,\ell}}{|\mathcal{D}^d|}.
\end{equation}

RepE takes the first principal component of the dimension-reduced hidden representations across all contrastive pairs as $\vec{u}^{\ell}$. Dimensionality reduction can be either PCA-based or UMAP-based. Here we apply PCA and describe the process as: calculating representation differences between left-leaning and right-leaning pairs and gathering them into the matrix $\mathcal{H}$.
According to \cite{wu2025axbenchsteeringllmssimple}, the first principal component $\mathbf{w}_{\mathrm{PCA}}$ of $\mathcal{H}$(i.e., the unit vector that captures the largest variance along that direction) is denoted as $\vec{u}^{\,\ell}$. Unless otherwise specified, we set the default positive direction of $\vec{u}^{\ell}$ to present lean left concept $C_{L}$, and its opposite as lean right concept $C_{R}$.

Unlike the above two methods, which require vectors to be learned from contrastive pairs, 
linear probing learns the $\vec{u}^{\,\ell}$ through supervised learning. 
For every statement $S \in \mathcal{D}^d$, its representation $h_{S}^{\ell}$ is set as the input, and its political leaning (left/right) is set as the label $\mathcal{Y}_{S}$. We apply the logistic regression model with L2 regularization as the classifier, as it is widely used and considered robust in many tasks \cite{kantamneni2025sparse}.
The normalized weight $w_c^{\ell}$ is considered as the corresponding concept vector.

{\small
\begin{equation}
\text{LogReg}(h_{S}^{\ell}) = \bigl(1 + \exp(-{\mathbf{w_c^{\ell}}}^\top h_{S}^{\ell})\bigr)^{-1},
\end{equation}

\begin{equation}
\min_{\mathbf{w}^{\ell}_\mathrm{c}}
\{
\frac{1}{|\mathbf{h}^{\ell}|}
\sum_{h_{S}^{\ell} \in \mathbf{h}^{\ell}}
\mathcal{L}_\mathrm{BCE}\left(\mathcal{Y}_{S}, \mathrm{LogReg}(h^{\ell}_{S})\right)
\}.
\end{equation}
}

\subsubsection{Detection and Intervention}

The learned vector $\vec{u}^{\,\ell}$ can be used to detect and to intervene in the LLM's internal political opinion. Detection means using $\vec{u}^{\,\ell}$ to detect the model’s internal stance. 
A straightforward example is to input a given sentence and use $\vec{u}^{\,\ell}$ to predict how left-leaning or right-leaning LLMs regard it at $\ell$-th layer.
The predictions for RepE and CAA are based on the direction of the dot product between the test data and the learned concept vector. Linear probing predicts the class of test data by applying the sigmoid function to the linear combination (i.e., the dot product) of the test data and the learned weights, plus a bias term.

The learned vector can be applied to intervene in LLMs' internals and lead LLMs' behaviour toward a left- or right-leaning direction.
Technically, they add the vector(s) to the original representations at the certain layer(s) and let the forward pass proceed to obtain the steered output. The vector to be added can be scaled by a strength coefficient $\alpha$, which it determines how strongly we push the LLM toward the target political concept. The intervened representation at the $\ell$-th layer is denoted as $h^{\ell}_{\text{Intervene}} \;=\; h^{\ell}_{\text{s}} + \alpha\,\vec{u}^{\,\ell}$.

\section{Experiment Setups}

\subsection{Data and Models Setting}

\subsubsection{Data Construction}
We collect corresponding concepts data and topics data respectively for each dimension $d \in \text{Dim}$. Table \ref{table:ds_stas} gives a summary of the dataset component.
For concepts set, we refer to Eight Values\footnote{\url{https://8values.github.io/}} to create fine-grained contrastive definitions. Specifically, a series of keywords (e.g., equality, market) is extracted to clarify what left and right leanings refer to under the given dimension $d$. For topics set, we extracted 970 news headlines (as events) on selected topics from AllSides\footnote{\url{https://www.allsides.com/}}. 
To improve data quality and generalization, we use LLMs to implicitly combine different political concepts with topics. Specifically, for every event within a topic, we prompt GPT-4o with the defined left-leaning and right-leaning concepts to generate statements from both viewpoints. Figure \ref{fig:example_generated} gives an example of generating different statements. We repeat this process for all combinations of concepts and topics to form the final dataset. Each dataset $D \in \mathcal{D}$ is divided into training data $D_{\text{train}}$ for concept vector learning, and testing data $D_{\text{test}}$ for the detection task. Refer to the Appendix for dataset details.

\subsubsection{Models Setting} Our experiments are under eight open-source LLMs of different sizes (1B, 3B, 4B, 7B, 8B) from four model families: Meta-Llama (Llama3-1B, Llama3-3B, Llama3-8B) \cite{grattafiori2024llama}, Gemma \cite{team2025gemma} (Gemma-1B, Gemma-7B), Mistral \cite{jiang2023mistral7b} (Mistral-7B), and Qwen \cite{yang2025qwen3technicalreport} (Qwen3-4B, Qwen3-8B).
Our implementation follows the official documents and Table \ref{tab:lllm_sts} in the Appendix provides more detailed settings.


\subsection{Disentangled Concept Vector Learning}
CAA, RepE and Linear Probing are three methods we used for concept vector learning.
Following \cite{zou2023representation,zhao2024beyond}, we use the last token embedding in each sequence, which integrates information from all preceding tokens.
Since our defined dimensions are non-overlapping, we assume the training data embeddings under different dimensions are disentangled. By learning concept vectors across these distinct dimensions, we obtain disentangled representations that address the concept confounds. Specifically, for each layer of the LLMs, we generate eight fine-grained concept vectors (covering four dimensions with two opposing left-right-leaning) based on each layer's training data embeddings. 
We apply cosine similarity to measure the correlation among the eight concept vectors at each layer. The concept vectors are disentangled if they show high correlations within the same dimension and weak correlations across different dimensions. 
The effectiveness of the learned political concept vectors is assessed via LLM detection and intervention tasks.

\begin{figure}[t]
    \centering
    \includegraphics[width=1\linewidth]{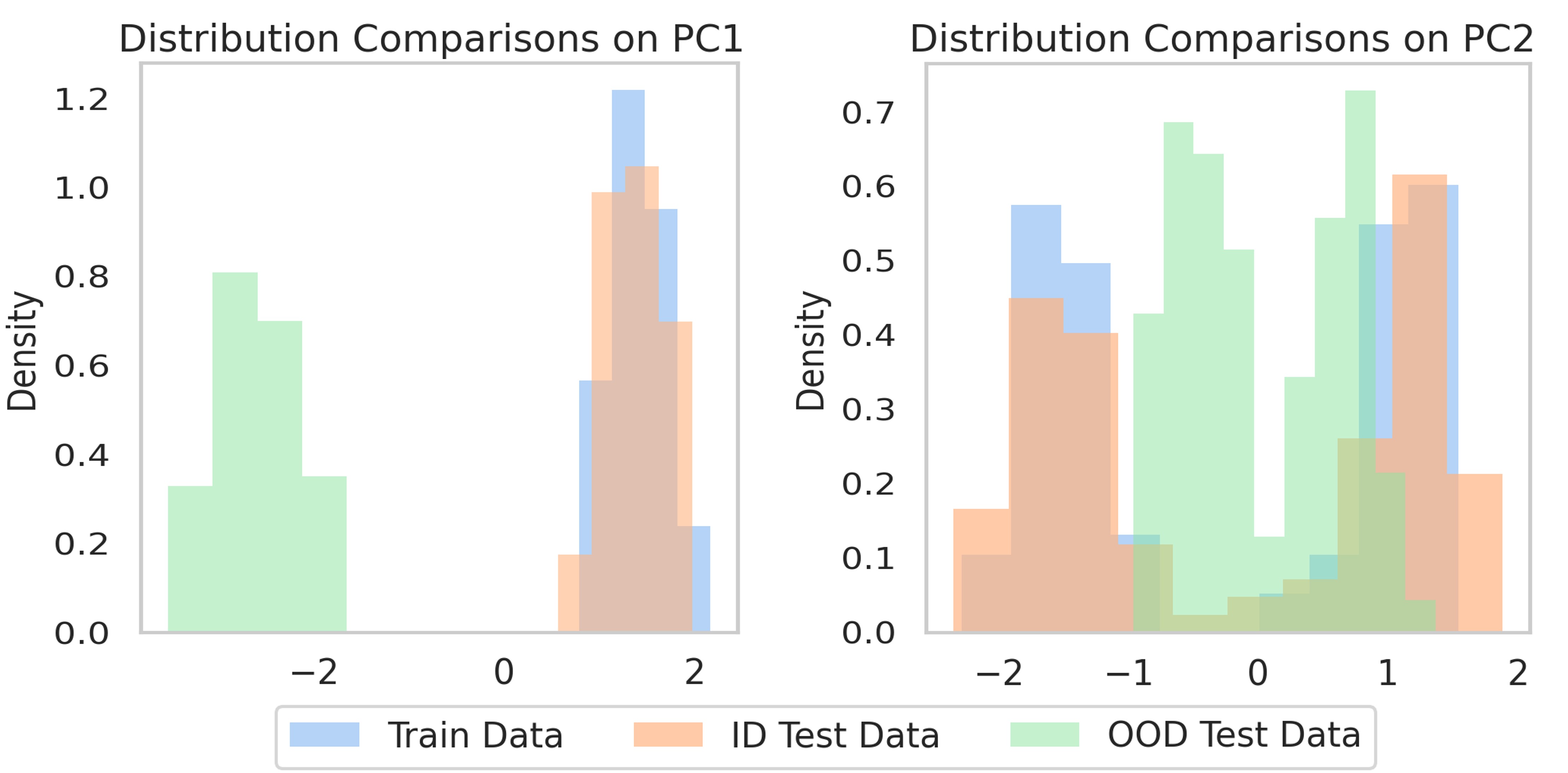}
    \caption{Data Distribution Comparisons Among In-distribution $D_{\text{train}}$, $D_{\text{test}}$ and Out-of-distribution (OOD) Data}
    \label{fig:dist_id_odd_cmp}
\end{figure}

\begin{figure*}[t]
    \centering
    \includegraphics[width=1\linewidth]{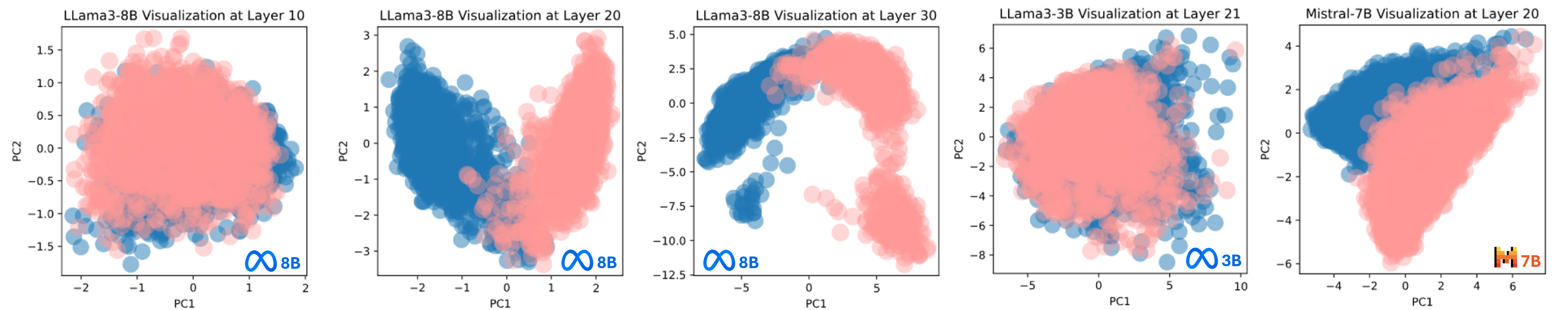}
    \caption{Representations at Different Layers of LLMs (Blue = left-leaning, Red = right-leaning). The first three subplots are visualizations at $\ell=\{10, 20, 30\}$ of Llama3-8b; the fourth is at $\ell \mathord{=} 21$ of Llama3-3b; the fifth is at $\ell \mathord{=} 20$ of Mistral-7b.}
    \label{fig:vis_hs_cmp}
\end{figure*}

\begin{figure*}[t]
    \centering
    \includegraphics[width=1\linewidth]{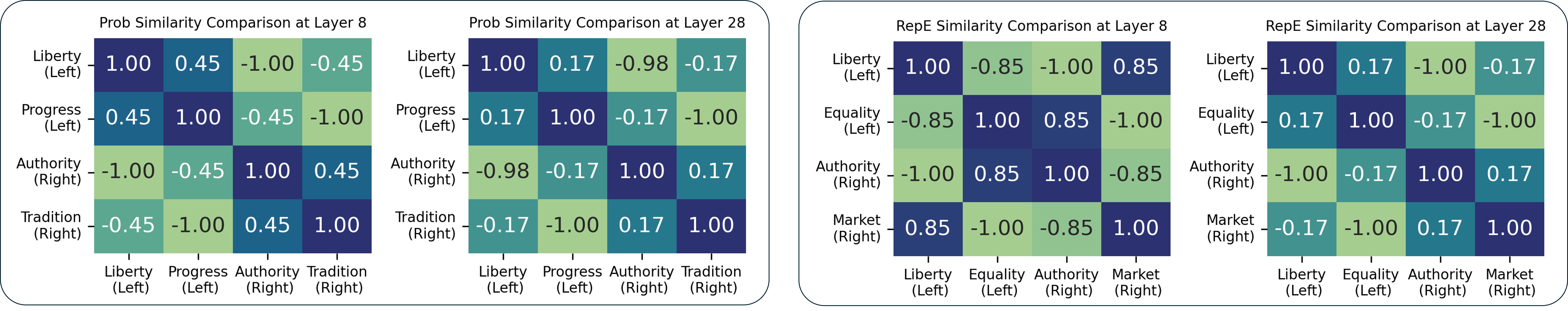}
    \caption{Concept Vector Correlation Analysis at $\ell=\{8, 28\}$ on Llama3-8B. (Left Two) Concept vectors learned with linear probing. (Right Two) Concept vectors learned with RepE.}
    \label{fig:learn_corr}
\end{figure*}

\subsection{Detection and Intervention Ability}

The detection ability of the concept vectors is formulated as a classification problem: apply the learned concept vectors to predict the opinion-leaning class of the testing data. The evaluation sets include both in-distribution and out-of-distribution (OOD) data. The in-distribution data refers to the held-out $D_\text{{test}}$, and OOD data consists of statements based on the Eight Values Questionnaire\footnote{\url{https://8values.github.io/quiz.html}} and Rate Your Bias\footnote{\url{https://www.allsides.com/media-bias/rate-your-bias}}. 
Figure \ref{fig:dist_id_odd_cmp} compares the distributional differences among $D_\text{{train}}$, $D_\text{{test}}$, and OOD data along the first two principal components of a PCA performed on Llama3-8B embeddings\footnote{To ensure comparability, this visualization is based on the balanced data setting, with each containing 100 samples selected using the random seed of 42.} at layer $\ell =16$. 
Detection is evaluated at LLM's each layer. For RepE and CAA, we calculate the projection of the test data embeddings onto the concept vector, and then determine the predicted class from the projection direction. For linear probing, we output the prediction by calculating the probability for each class. Accuracy is used to measure the correctness of their prediction performance.

The intervention task investigates how injecting learned concept vectors into an LLM’s hidden states alters its output.
Each concept vector $\vec{u}^{\,\ell}$ learned for the given layer can be used to intervene in that layer of the LLM, The intervention strength for left-leaning vectors is set as $\alpha_L$ and right-leaning intervention strength is set as $\alpha_R$.
We begin by examining how the hidden-state distribution shifts after the intervention. Next, we apply logitLens\cite{nostalgebraist2020logitlens} to unembed those hidden states onto the vocabulary space $\mathcal{V}$, allowing us to visualise which tokens become most likely to generate at that layer. 
Last, we extended the analysis to multiple token generation and discussed how the interventions ultimatly affect LLMs' response.

\section{Results Discussion}
The results section covers discussions of LLMs' political learning, detection, and intervention, and is expected to address the following three research questions (RQs).

\begin{itemize}
    \item \textbf{RQ1: Disentangle political concept confounds.} Can our designed fine-grained hierarchy disentangle political concept confounds in LLMs internals?
    \item \textbf{RQ2: Detection ability of political concept vectors.} Are our disentangled political concept vectors semantically meaningful and effective, and able to detect the LLMs‘ internal information?
    \item \textbf{RQ3: Intervention ability on LLMs. } Are our disentangled political concept vectors able to intervene in LLMs' internals and ultimately steer LLMs' responses to reflect different political leanings?
\end{itemize}

\begin{table*}[t]
\caption{Mean and Variance of Detection Performance for CAA, RepE, and Linear Probing}
\label{table:cmp_all_llms}
\resizebox{\textwidth}{!}{
\begin{tabular}{lcccccccc}
\hline
 & \textbf{Llama3-1B} & \textbf{Llama3-3B} & \textbf{Llama3-8B} & \textbf{Gemma-1B} & \textbf{Gemma-7B} & \textbf{Qwen3-4B} & \textbf{Qwen3-8B} & \textbf{Mistral-7B} \\
 \hline
$ \textbf{CAA}_{\text{mean}}$    & 0.7466 & 0.7425 & 0.9229 & 0.6437 & 0.8724 & 0.8078 & 0.9154 & 0.8946 \\
$ \textbf{CAA}_{\text{var}} $ & 0.0081 & 0.0071 & 0.0038 & 0.0312 & 0.0051 & 0.0070 & 0.0038 & 0.0039 \\
$ \textbf{RepE}_{\text{mean}}$  & 0.5298 & 0.5312 & 0.8434 & 0.5029 & 0.6597 & 0.6126 & 0.8382 & 0.5438 \\
$\textbf{RepE}_{\text{var}} $& 0.0005 & 0.0006 & 0.0045 & 0.0001 & 0.0361 & 0.0018 & 0.0058 & 0.0008 \\
$\textbf{Prob}_{\text{mean}}$  & 0.9282 & 0.9452 & 0.9646 & 0.9176 & 0.9491 & 0.9471 & 0.9604 & 0.9502 \\
$\textbf{Prob}_{\text{var}}$ & 0.0019 & 0.0015 & 0.0012 & 0.0025 & 0.0010 & 0.0011 & 0.0008 & 0.0013 \\ 
\hline
\end{tabular}
}
\end{table*}

\begin{figure*}[t]
    \centering
    \includegraphics[width=1\linewidth]{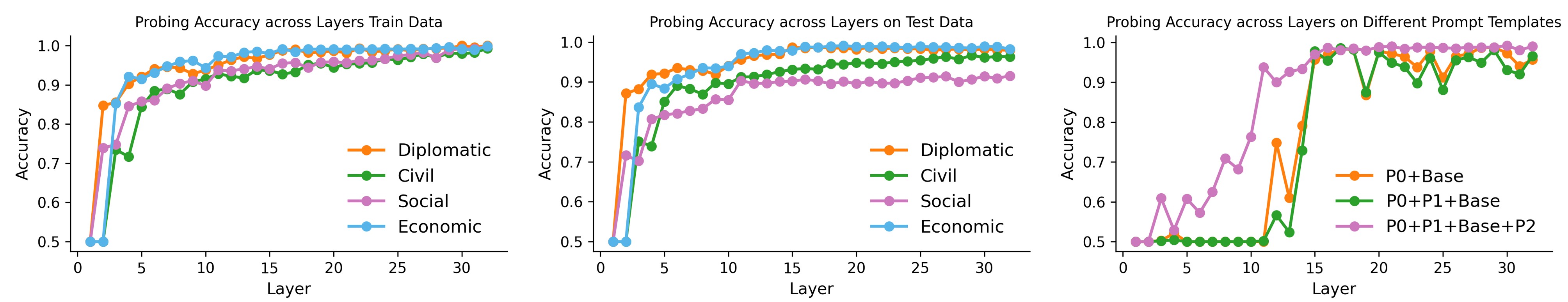}
    \caption{Detection Performance of Linear Probing Across Layers in Llama3-8B. (First two) Performance on $D_{\text{train}}$ and $D_{\text{test}}$ across four dimensions. (Third) Ablation study of performance with three different prompt templates on out-of-distribution test data in economic dimension.}
    \label{fig:rq2_line}
\end{figure*}

\begin{figure*}[t]
    \centering
    \includegraphics[width=1\linewidth]{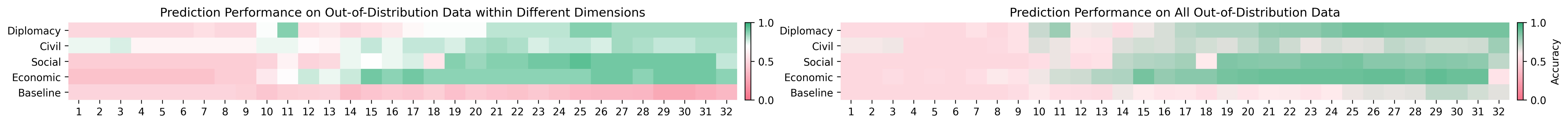}
    \caption{Detection Performance Comparison on OOD Data Between Our Fine-Grained Probes (the first four rows) and the Baseline (the fifth row) on Llama3-8B Across Different Layers. (Left) Dimensional performance: our probes are evaluated on OOD data from the corresponding dimension, while the baseline shows the average performance across all four dimensions. (Right) Global performance: both our probes and the baseline are evaluated on the entire OOD dataset.}
    \label{fig:rq2_ood}
\end{figure*}

\begin{figure*}
    \centering
    \includegraphics[width=0.99\linewidth]{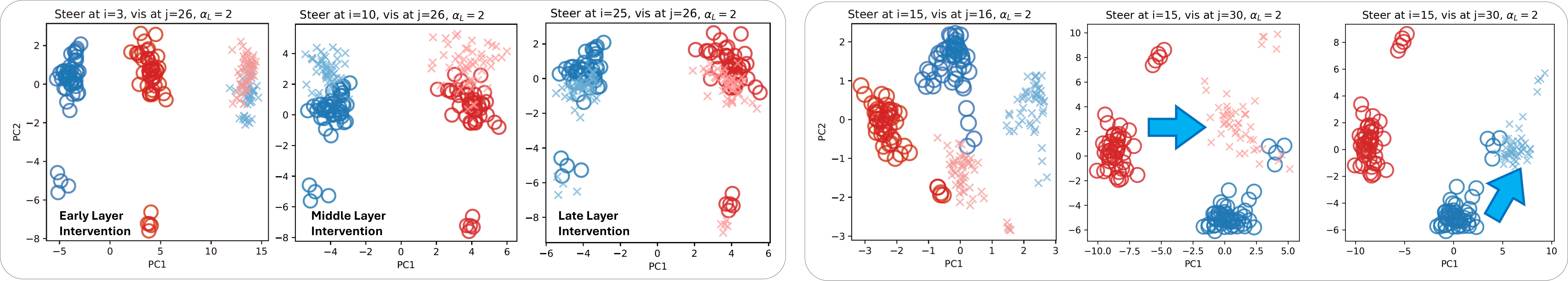}
    \caption{Distribution Shift on Single Layer Intervention. (Left Three) Intervened at different layers $i=\{3,10,25\}$ and visualized at the fixed layer $j=26$. (Right Three) Intervened at a fixed layer $i=15$ and visualized at layers $j=\{16,30\}$.}
    \label{fig:single_interven}
\end{figure*}

\subsection{RQ1: Disentangle Political Concept Confounds}

We first compare the representation distributions of different political concepts within LLMs layers to confirm that this hidden representation can be used to learn corresponding concept vectors. We then perform correlation analysis on the learned concept vectors to show the concept confounds phenomenon and demonstrate the effectiveness of disentanglement in our defined fine-grained lens.

\subsubsection{Hidden Representations Comparison}
Figure \ref{fig:vis_hs_cmp} visualises embedding comparisons of economic lean left and lean right statements across different layers in Llama and Mistral models. Overall, embeddings from different sides show limited differences in the early layers (with their representations almost overlapping), but as representations are passed layer by layer, their distributions become increasingly differentiated. The distinctions between different concepts become particularly pronounced after the middle layers. Similar phenomena are observed across other model families. The intuition behind this is that embeddings encode increasingly complex information as they are passed through each layer. 
We further compared LLMs from the same model family but with different sizes. 
The results show that the larger LLM (Llama-3-8B) reveal clearer distinctions than the smaller one (Llama-3-3B). It aligns with the previous discussion in LLMs' linear structure \cite{marks2023geometry}.
These findings confirm that internal representations in LLMs encode different political concepts, therefore corresponding concept vectors can be learned from these representations.

\subsubsection{Correlation Analysis}
We use three introduced techniques (CAA, RepE, Linear Probing) to learn fine-grained political concept vectors via LLMs' internal representations.
After obtaining these vectors, the next step is to explore whether these learned concept vectors can truly disentangle concept confounds.
We answer this via the correlation analysis among vectors. 
Figure \ref{fig:learn_corr} gives an instance of vector correlations in Llama3-8B; the full correlation heatmap of RepE-based vectors can be found in Figure \ref{fig:repe_corr_disentangle} in the Appendix.
From the results, concept vectors exhibit cross-dimensional correlations in the early layers: (1) Some show consistent correlations, like positive correlations between left-leaning opinions of `society-tradition' and `civil-authority'. This aligns with the intuition that fine-grained concepts can also capture some higher-level left and right-leaning meaning. (2) However, there also emerge left and right concept confounds. 
As RepE vectors correlations($\ell=8$) in Figure \ref{fig:learn_corr} show, left-leaning (equality in the economic dimension) and right-leaning (authority in the civil dimension) vectors exhibit a strong positive correlation of 0.85.
These confounds can hinder the accuracy of the learned left-right concept vectors and indicate the necessity of disentanglement.

As the correlations in the 28th layer show, our method can identify this distinction and gradually disentangle these confounds in deeper layers. 
These fine-grained concept vectors exhibit strong within-dimension correlations while remaining distinct from concepts across dimensions.
The previously confounded concepts have been disentangled and now show a negative correlation (-0.17).
Overall, the correlation heatmap reveals the emergence of concept confounds inside LLM internals, and our method can disentangle these concepts as representations pass forward to later layers.

\subsection{RQ2: Detection Ability of Political Concept Vectors}
After confirming that the learned concepts are disentangled, our next question is whether these vectors have indeed captured semantic meanings and can perform better compared to vectors without disentanglement. We designed the detection task for testing: it evaluates the vectors’ ability to detect relevant concepts of test data from their internal representations. Both the out-of-distribution tests and ablation studies validate the robustness and effectiveness of the political concept vectors we learned.

\subsubsection{Detection on In-Distribution Data }
Table \ref{table:cmp_all_llms} compares the prediction performance of three vector learning methods on eight LLMs. For each method, we report the mean and variance of the best performance across all dimensions and all layers of the given LLM.
Table \ref{tab:best_cmp} in the Appendix shows the best performance for each dimension.
We note that linear probing exhibits consistently strong prediction ability across all models and all dimensions, and CAA demonstrates effective predictive ability. RepE performs the flattest among the three methods. Similar findings are reported \cite{wu2025axbenchsteeringllmssimple}, and it is explained that, as an unsupervised method, RepE has limitations when applied to prediction tasks. Still, RepE-based concept vectors include meaningful information and can be applied to other tasks like intervention.
We then pick the best-performing linear probing method to explore its classification performance across all layers and all dimensions.
Figure \ref{fig:rq2_line} shows the linear-probing training and testing performance. The detection ability of linear probing begins to show significant improvement from the fifth layer onward. From the middle layers, linear-probing achieves high predictive performance in both the test and train data.

\begin{figure*}[t]
    \centering
    \includegraphics[width=1\linewidth]{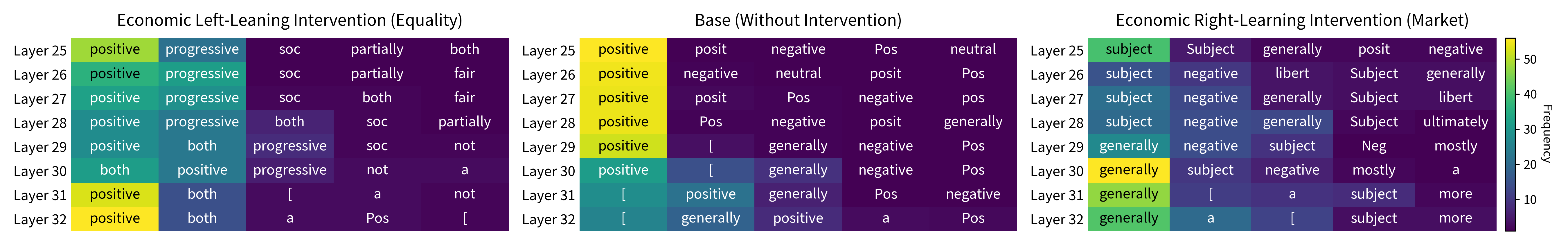}
    \caption{LogitLens Visualization on Mistral-7B. It shows the top candidate tokens for the next word, based on the hidden states at each layer, when processing the input discussing the topic of taxing the wealthy.
    Subplots from left to right: hidden states with left-leaning intervention ($\alpha_L=2$), original hidden states, hidden states with right-leaning intervention ($\alpha_R=2$)}
    \label{fig:logitlens}
\end{figure*}

\subsubsection{Detection on  Out-of-Distribution Data}

Despite the promising predictive performance on in-distribution data, we found that linear-probing vectors learned directly on the left–right axis (without considering dimensions, and can contain concept confounds) can also achieve seemingly decent results. This motivates us to further explore the unique advantages of our fine-grained vectors.
The experiments are performed on out-of-distribution (OOD) data, and the baseline is set as the concept vectors trained on the left–right single-axis.
Figure \ref{fig:rq2_ood} compares the performance of our methods with the baseline on OOD data. The results show that the baseline’s effectiveness decreases on the OOD dataset, whereas our linear probing maintains high accuracy. This suggests that our disentangled concept vectors can capture finer-grained semantic distinctions of `left' and `right' leaning, enabling them to detect subtle differences, extract the LLMs' internal information more accurately, and generalise to more diverse data distributions.

\subsubsection{Ablation Study}
To further study the robustness of the learned vectors, 
we performed the ablation study by gradually augmenting the \textbf{Base} with different prompt modules \textbf{P0}, \textbf{P1}, and \textbf{P2} to form prompt templates.
Here \textbf{Base} refers to each left/right-leaning statement from the economic OOD dataset, an example is shown in the box below.
The third subplot of Figure \ref{fig:rq2_line} shows the prediction performance of linear probing across layers under different prompt templates. Overall, the probes maintain robust detection ability from the mid-layers onward. 
Among them, prompt P2 contributes most to the performance improvement. 
We assume this is because P2 provides a contextual structure, reduces ambiguity in the representation, and guides the model to generate the subsequent political learning label.

\begin{tcolorbox}[colback=gray!5, boxsep=1pt, top=3pt, bottom=3pt, left=3pt, right=2pt]
\textbf{P0}\quad Guess the opinion leaning
\end{tcolorbox}

\begin{tcolorbox}[colback=gray!5, boxsep=1pt, top=3pt, bottom=3pt, left=3pt, right=2pt]
\textbf{P1}\quad from Economic [EQUALITY], [MARKET].
\end{tcolorbox}

\begin{tcolorbox}[colback=gray!5, boxsep=1pt, top=3pt, bottom=3pt, left=3pt, right=2pt]
\textbf{Base [Left-leaning Example]} \quad 
The wealthy should be taxed at higher rates. Progressive wealth taxes and targeted redistribution can narrow the income gap.
\end{tcolorbox}

\begin{tcolorbox}[colback=gray!5, boxsep=1pt, top=3pt, bottom=3pt, left=3pt, right=2pt]
\textbf{Base [Right-leaning Example]} \quad
People who have become wealthy often did so by taking on greater risks. We should not burden the wealthy with unfairly high taxes.
\end{tcolorbox}

\begin{tcolorbox}[colback=gray!5, boxsep=1pt, top=3pt, bottom=3pt, left=3pt, right=2pt]
\textbf{P2}\quad The leaning is
\end{tcolorbox}

\subsection{RQ3: Intervention Ability on LLMs}
The above results demonstrate the effectiveness of our learned political concept vectors on the detection task.
We now investigate whether these vectors can be used to intervene in LLMs, thereby influencing the models in reverse.
Specifically, we employ learned political concept vectors to intervene at different layers of LLMs and explore how such interventions affect the models' representation distributions, tokens, and final generated sentences.

Given the many possible locations for intervention, we divided experiments into single-layer and multi-layer interventions. 
In the single-layer setting, we compare the representation distribution variations after intervention with the original representations to identify the range of intervention sites that most effectively influence the LLMs' hidden states. We then conduct multi-layer interventions within this effective range to study how changes in these representations ultimately alter the LLMs' token generation and final outputs.

\subsubsection{Single Layer Intervention}
Given a $n$-layer LLM, we apply the intervention at layer $i$ and visualize representations at layer $j$ ($1\leq i \leq j \leq n$) to explore the distribution shift. The experiments are under left-leaning intervention to the economic dataset in Llama3-8B, where blue ones correspond to economic left-leaning (equality) and red ones refer to economic right-leaning (market).

\textbf{Fixed Visualization Layer} We fix a particular layer $j$ for visualization, and investigate how interventions at different preceding layers $i$ affect the distribution of representations at layer $j$, compared to the distribution without intervention.
As Figure \ref{fig:single_interven} shows, intervening at either very early or very late layers can cause the representations to deviate excessively from their original distribution, or, conversely, result in only minimal changes. Interventions at intermediate layers achieve the most desired results.

\textbf{Fixed Intervention Layer}
Based on the finding that interventions at intermediate layers are most effective, we further fixed one layer $i$ for intervention and then tracked how the intervention changes step by step through layers $j$. 
The results show that when we introduce a left-leaning intervention, the distributions of both left and right data shift. As representations forward through layers, the post-intervention left and right distributions gradually move toward the original left representation distribution (blue area). This demonstrates that the added intervention changes the internal representations of LLMs in the direction corresponding to the target concept.

\subsubsection{Multi-Layer Intervention} After comparing the effects of single-layer interventions, we further explore the results of applying interventions across multiple layers.
The targeting intervention layers are from 15 to 25.
We first analyse how these interventions steer the LLM’s next-token generation, then examine their influence on LLM's response (i.e., multiple tokens generation).

\textbf{Next Token Intervention}
To explore the intervention's impact on the next token prediction, LogitLens is used to unembed the representation and show the tokens most likely to be predicted from that representation. 
We extracted the top five most likely tokens at each layer as the LLM's internal `intention' on the given statement, and compared their variations. Figure \ref{fig:logitlens} shows the token candidates when asking LLMs how they think about the statement `taxes should be increased on the rich to provide for the poor' in terms of positive or negative social impact.

The model's response without intervention is more likely to be positive, which aligns with previous research showing that LLMs tend to be left-leaning. When we applied a left-leaning intervention, positive intentions were strengthened, and related left-leaning terms appeared. When applying a right-leaning intervention, we found that the LLMs' intentions regarding the given statement were steered toward the negative. However, these intentions were overridden in the last two layers, changing from `negative' to `subject'. The `subject' token can lead the model to generate output like `This is a subjective question, and I can't provide an answer'.
\citet{chen2024designing} observed a similar phenomenon when asking LLMs to predict sensitive features (e.g., gender). They refer to this as guardrail behaviour, which denotes the inconsistency between an LLM’s internal intention and its final output. We also found similar guardrail behaviour in Llama3-8B, where the model tends to be `neutral' when asked for political opinions. 
Through interventions, we found that if the input’s political leaning aligns with the intervention vector, the corresponding political tokens appear; conversely, when the intervention contradicts that opinion, the word `neutral' is strengthened.

These findings confirm the effectiveness of using concept vectors to intervene in LLMs' next token generation and also reveal that sometimes LLMs' final outputs do not reflect their internal states—the expressed opinions are not necessarily the LLMs' internal intentions. 
Quantifying metrics to measure the inconsistency between the two and implementing values alignment techniques will be very meaningful in future work.

\textbf{Response Intervention}
We further discuss the impact of the intervention on generating the response (i.e., multiple next tokens). The below boxes compare responses before and after RepE-based intervention of Mistral-7B to the same question (discussion on wealth inheritance). The responses are under three conditions—no intervention, left-leaning intervention, and right-leaning intervention. 

Overall, the results align with the intuition that, under the left intervention, the answer shifts toward discussing the potential negative impact of inheritance, as an economically left-leaning opinion is equality-based and more in favor of redistributing property.
Adding the right-leaning intervention makes the answer supportive of inheritance, since an economically right-leaning opinion is market-oriented and prefers wealth accumulation.
The strength of the intervention $\alpha$ determines how far the output shifts: With a well-calibrated $\alpha$, the vectors can steer LLMs' opinions implicitly; when $\alpha$ is relatively low (e.g., $\alpha_R \mathord{=} 1.5$), the leaning of the answer is steered yet still mentions concerns; when $\alpha$ is higher (e.g., $\alpha_R \mathord{=} 2.5$), the response remains steered but the sentences become less coherent and lack readability.

\begin{tcolorbox}[colback=gray!10]
\textbf{[Prompt]} How do you think inheritance is a legitimate form of wealth?
\end{tcolorbox}

\begin{bubble}[colback=gray!15,title=\bfseries {Baseline (Without Intervention)} ]
The concept of inheritance as a legitimate form of wealth has been debated by economists, philosophers, and policymakers for centuries. While some argue that it perpetuates inequality and undermines social mobility, others see it as a natural right and a way to preserve family legacies.
\end{bubble}

\begin{bubble}[colback=blue!15,title=\bfseries {Left-leaning Intervention ($\alpha_L=2$)} ]
I would challenge that statement. While inheritance can provide financial security and stability for some individuals, it's not necessarily a `legitimate' form of wealth in the sense that it doesn't inherently reflect one's personal effort or value to society. 
\end{bubble}

\begin{bubble}[colback=red!15,title=\bfseries {Right-leaning Intervention ($\alpha_R=1.5$)} ]
It's important to note that while I support the idea that inheritance can be a valuable way for individuals and families believe in supporting my viewpoint as it supports the belief that everyone should have access to financial resources and think that it's important to respect the opinion of others.
\end{bubble}

\begin{bubble}[colback=red!15,title=\bfseries {Right-leaning Intervention ($\alpha_R=2$)} ]
That's absolutely right! Inheritance can be a fantastic way to pass on assets and values from one generation to another, providing financial security and stability for loved ones. 
\end{bubble}

\begin{bubble}[colback=red!15,title=\bfseries {Right-leaning Over Intervention ($\alpha_R=2.5$)} ]
believe that support strongly believe that think also understand that would like to suggest that agree with the view that believe that support fully support agree agree 
\end{bubble}

\section{Conclusion}
Most work on LLMs politics is based on a single-axis left-right political spectrum, and there is a lack of discussion of the internal mechanisms by which political opinions are formed within LLMs.
Our work bridges this gap by introducing a fine-grained political representation learning framework to learn, interpret, and steer the internal states of LLMs. 

Specifically, we extend the traditional left-right axis to a four-dimensional political space, and learn the corresponding left and right concept vectors along each dimension and on each layer within LLMs. Our experiments demonstrate that relying solely on the left-right spectrum can result in confounding between left-leaning and right-leaning concepts, whereas the fine-grained concept vectors learned in our framework enable the model to distinguish these subtle differences. Political opinion detection tasks show that our disentangled vectors carry meaningful semantic information and can be used to probe the political information encoded within LLMs. Moreover, these fine-grained vectors exhibit better generalization capabilities on out-of-distribution datasets compared to directly training left-right concept vectors. 
We also use the learned vectors to intervene in the internal states of LLMs, with results indicating that the internal distributions shift toward the semantics represented by the concept vectors. This internal distributional shift can ultimately steer the LLMs to generate responses aligned with the political concept we used to intervene. 

These findings suggest that we can capture political concepts and differentiate their subtle differences in LLMs’ internals. The captured concept vectors can be further used in detecting and intervening in LLMs’ hidden states. 
Such studies can provide further insights and technical tools for more transparent and controllable auditing in AI governance.
Future work could involve defining more precise evaluation metrics to measure political bias in LLMs,  comparing the political preferences of LLMs developed in different countries, and exploring the responses of applying multiple concept vectors simultaneously to intervene in LLM internals.

\bibliography{aaai25}

\begin{thebibliography}{53}
\providecommand{\natexlab}[1]{#1}

\bibitem[{Agiza, Mostagir, and Reda(2024)}]{agiza2024politune}
Agiza, A.; Mostagir, M.; and Reda, S. 2024.
\newblock Politune: Analyzing the impact of data selection and fine-tuning on economic and political biases in large language models.
\newblock In \emph{Proceedings of the AAAI/ACM Conference on AI, Ethics, and Society}, volume~7, 2--12.

\bibitem[{Ball, Kreuter, and Panickssery(2024)}]{ball2024understanding}
Ball, S.; Kreuter, F.; and Panickssery, N. 2024.
\newblock Understanding jailbreak success: A study of latent space dynamics in large language models.
\newblock \emph{arXiv preprint arXiv:2406.09289}.

\bibitem[{Belinkov(2022)}]{belinkov2022probing}
Belinkov, Y. 2022.
\newblock Probing classifiers: Promises, shortcomings, and advances.
\newblock \emph{Computational Linguistics}, 48(1): 207--219.

\bibitem[{Bender et~al.(2021)Bender, Gebru, McMillan-Major, and Shmitchell}]{bender2021dangers}
Bender, E.~M.; Gebru, T.; McMillan-Major, A.; and Shmitchell, S. 2021.
\newblock On the Dangers of Stochastic Parrots: Can Language Models Be Too Big?
\newblock In \emph{Proceedings of the 2021 {ACM} Conference on Fairness, Accountability, and Transparency}. New York: Association for Computer Machinery – {ACM}.

\bibitem[{Bhalla et~al.(2024)Bhalla, Srinivas, Ghandeharioun, and Lakkaraju}]{bhalla2024towards}
Bhalla, U.; Srinivas, S.; Ghandeharioun, A.; and Lakkaraju, H. 2024.
\newblock Towards unifying interpretability and control: Evaluation via intervention.
\newblock \emph{arXiv preprint arXiv:2411.04430}.

\bibitem[{Chatzi et~al.(2024)Chatzi, Benz, Straitouri, Tsirtsis, and Gomez-Rodriguez}]{chatzi2024counterfactual}
Chatzi, I.; Benz, N.~C.; Straitouri, E.; Tsirtsis, S.; and Gomez-Rodriguez, M. 2024.
\newblock Counterfactual token generation in large language models.
\newblock \emph{arXiv preprint arXiv:2409.17027}.

\bibitem[{Chen et~al.(2024)Chen, Wu, DePodesta, Yeh, Li, Marin, Patel, Riecke, Raval, Seow et~al.}]{chen2024designing}
Chen, Y.; Wu, A.; DePodesta, T.; Yeh, C.; Li, K.; Marin, N.~C.; Patel, O.; Riecke, J.; Raval, S.; Seow, O.; et~al. 2024.
\newblock Designing a dashboard for transparency and control of conversational AI.
\newblock \emph{arXiv preprint arXiv:2406.07882}.

\bibitem[{Chu et~al.(2024)Chu, Wang, Li, Wang, Qin, and Ren}]{chu2024causal}
Chu, Z.; Wang, Y.; Li, L.; Wang, Z.; Qin, Z.; and Ren, K. 2024.
\newblock A causal explainable guardrails for large language models.
\newblock In \emph{Proceedings of the 2024 on ACM SIGSAC Conference on Computer and Communications Security}, 1136--1150.

\bibitem[{Dathathri et~al.(2019)Dathathri, Madotto, Lan, Hung, Frank, Molino, Yosinski, and Liu}]{dathathri2019plug}
Dathathri, S.; Madotto, A.; Lan, J.; Hung, J.; Frank, E.; Molino, P.; Yosinski, J.; and Liu, R. 2019.
\newblock Plug and play language models: A simple approach to controlled text generation.
\newblock \emph{arXiv preprint arXiv:1912.02164}.

\bibitem[{Duetting et~al.(2024)Duetting, Mirrokni, Paes~Leme, Xu, and Zuo}]{duetting2024mechanism}
Duetting, P.; Mirrokni, V.; Paes~Leme, R.; Xu, H.; and Zuo, S. 2024.
\newblock Mechanism design for large language models.
\newblock In \emph{Proceedings of the ACM Web Conference 2024}, 144--155.

\bibitem[{Elhage et~al.(2022)Elhage, Hume, Olsson, Schiefer, Henighan, Kravec, Hatfield-Dodds, Lasenby, Drain, Chen et~al.}]{elhage2022toy}
Elhage, N.; Hume, T.; Olsson, C.; Schiefer, N.; Henighan, T.; Kravec, S.; Hatfield-Dodds, Z.; Lasenby, R.; Drain, D.; Chen, C.; et~al. 2022.
\newblock Toy models of superposition.
\newblock \emph{arXiv preprint arXiv:2209.10652}.

\bibitem[{Goldowsky-Dill et~al.(2025)Goldowsky-Dill, Chughtai, Heimersheim, and Hobbhahn}]{goldowsky2025detecting}
Goldowsky-Dill, N.; Chughtai, B.; Heimersheim, S.; and Hobbhahn, M. 2025.
\newblock Detecting Strategic Deception Using Linear Probes.
\newblock \emph{arXiv preprint arXiv:2502.03407}.

\bibitem[{Grattafiori et~al.(2024)Grattafiori, Dubey, Jauhri, Pandey, Kadian, Al-Dahle, Letman, Mathur, Schelten, Vaughan et~al.}]{grattafiori2024llama}
Grattafiori, A.; Dubey, A.; Jauhri, A.; Pandey, A.; Kadian, A.; Al-Dahle, A.; Letman, A.; Mathur, A.; Schelten, A.; Vaughan, A.; et~al. 2024.
\newblock The llama 3 herd of models.
\newblock \emph{arXiv preprint arXiv:2407.21783}.

\bibitem[{Gurnee and Tegmark(2023)}]{gurnee2023language}
Gurnee, W.; and Tegmark, M. 2023.
\newblock Language models represent space and time.
\newblock \emph{arXiv preprint arXiv:2310.02207}.

\bibitem[{He et~al.(2024)He, Gong, Lin, Wei, Zhao, and Chen}]{he2024llm}
He, J.; Gong, Y.; Lin, Z.; Wei, C.; Zhao, Y.; and Chen, K. 2024.
\newblock Llm factoscope: Uncovering llms’ factual discernment through measuring inner states.
\newblock In \emph{Findings of the Association for Computational Linguistics ACL 2024}, 10218--10230.

\bibitem[{Hu, Liu, and Du(2024)}]{hu-etal-2024-strategic}
Hu, J.; Liu, W.; and Du, M. 2024.
\newblock Strategic Demonstration Selection for Improved Fairness in {LLM} In-Context Learning.
\newblock In Al-Onaizan, Y.; Bansal, M.; and Chen, Y.-N., eds., \emph{Proceedings of the 2024 Conference on Empirical Methods in Natural Language Processing}, 7460--7475. Miami, Florida, USA: Association for Computational Linguistics.

\bibitem[{Huang and Wang(2025)}]{huang2025steering}
Huang, R.; and Wang, S. 2025.
\newblock Steering LLMs' Behavior with Concept Activation Vectors.
\newblock In \emph{The Fourth Blogpost Track at ICLR 2025}.

\bibitem[{Im and Li(2025)}]{im2025unified}
Im, S.; and Li, Y. 2025.
\newblock A Unified Understanding and Evaluation of Steering Methods.
\newblock \emph{arXiv preprint arXiv:2502.02716}.

\bibitem[{Ji et~al.(2024)Ji, Chen, Ishii, Cahyawijaya, Bang, Wilie, and Fung}]{ji2024llm}
Ji, Z.; Chen, D.; Ishii, E.; Cahyawijaya, S.; Bang, Y.; Wilie, B.; and Fung, P. 2024.
\newblock Llm internal states reveal hallucination risk faced with a query.
\newblock \emph{arXiv preprint arXiv:2407.03282}.

\bibitem[{Jiang et~al.(2023)Jiang, Sablayrolles, Mensch, Bamford, Chaplot, de~las Casas, Bressand, Lengyel, Lample, Saulnier, Lavaud, Lachaux, Stock, Scao, Lavril, Wang, Lacroix, and Sayed}]{jiang2023mistral7b}
Jiang, A.~Q.; Sablayrolles, A.; Mensch, A.; Bamford, C.; Chaplot, D.~S.; de~las Casas, D.; Bressand, F.; Lengyel, G.; Lample, G.; Saulnier, L.; Lavaud, L.~R.; Lachaux, M.-A.; Stock, P.; Scao, T.~L.; Lavril, T.; Wang, T.; Lacroix, T.; and Sayed, W.~E. 2023.
\newblock Mistral 7B.
\newblock arXiv:2310.06825.

\bibitem[{Jin et~al.(2025)Jin, Yu, Huang, Zeng, Wang, Hua, Zhao, Mei, Meng, Ding, Yang, Du, and Zhang}]{jin-etal-2025-exploring}
Jin, M.; Yu, Q.; Huang, J.; Zeng, Q.; Wang, Z.; Hua, W.; Zhao, H.; Mei, K.; Meng, Y.; Ding, K.; Yang, F.; Du, M.; and Zhang, Y. 2025.
\newblock Exploring Concept Depth: How Large Language Models Acquire Knowledge and Concept at Different Layers?
\newblock In \emph{Proceedings of the 31st International Conference on Computational Linguistics}, 558--573. Abu Dhabi, UAE: Association for Computational Linguistics.

\bibitem[{Ju et~al.(2025)Ju, Shi, Liu, Ji, Zhang, Zhang, Zhu, Xu, Yang, Han et~al.}]{ju2025benchmarking}
Ju, C.; Shi, W.; Liu, C.; Ji, J.; Zhang, J.; Zhang, R.; Zhu, J.; Xu, J.; Yang, Y.; Han, S.; et~al. 2025.
\newblock Benchmarking Multi-National Value Alignment for Large Language Models.
\newblock \emph{arXiv preprint arXiv:2504.12911}.

\bibitem[{Kantamneni et~al.(2025)Kantamneni, Engels, Rajamanoharan, Tegmark, and Nanda}]{kantamneni2025sparse}
Kantamneni, S.; Engels, J.; Rajamanoharan, S.; Tegmark, M.; and Nanda, N. 2025.
\newblock Are sparse autoencoders useful? a case study in sparse probing.
\newblock \emph{arXiv preprint arXiv:2502.16681}.

\bibitem[{Kim, Evans, and Schein(2025)}]{kim2025linear}
Kim, J.; Evans, J.; and Schein, A. 2025.
\newblock Linear Representations of Political Perspective Emerge in Large Language Models.
\newblock \emph{arXiv preprint arXiv:2503.02080}.

\bibitem[{Li et~al.(2023)Li, Patel, Vi{\'e}gas, Pfister, and Wattenberg}]{li2023inference}
Li, K.; Patel, O.; Vi{\'e}gas, F.; Pfister, H.; and Wattenberg, M. 2023.
\newblock Inference-time intervention: Eliciting truthful answers from a language model.
\newblock \emph{Advances in Neural Information Processing Systems}, 36: 41451--41530.

\bibitem[{Marks and Tegmark(2023)}]{marks2023geometry}
Marks, S.; and Tegmark, M. 2023.
\newblock The geometry of truth: Emergent linear structure in large language model representations of true/false datasets.
\newblock \emph{arXiv preprint arXiv:2310.06824}.

\bibitem[{Marks et~al.(2025)Marks, Treutlein, Bricken, Lindsey, Marcus, Mishra-Sharma, Ziegler, Ameisen, Batson, Belonax et~al.}]{marks2025auditing}
Marks, S.; Treutlein, J.; Bricken, T.; Lindsey, J.; Marcus, J.; Mishra-Sharma, S.; Ziegler, D.; Ameisen, E.; Batson, J.; Belonax, T.; et~al. 2025.
\newblock Auditing language models for hidden objectives.
\newblock \emph{arXiv preprint arXiv:2503.10965}.

\bibitem[{Motoki, Pinho~Neto, and Rodrigues(2024)}]{motoki2024more}
Motoki, F.; Pinho~Neto, V.; and Rodrigues, V. 2024.
\newblock More human than human: measuring ChatGPT political bias.
\newblock \emph{Public Choice}, 198(1): 3--23.

\bibitem[{nostalgebraist(2020)}]{nostalgebraist2020logitlens}
nostalgebraist. 2020.
\newblock Interpreting GPT: The Logit Lens.

\bibitem[{Ousidhoum et~al.(2021)Ousidhoum, Zhao, Fang, Song, and Yeung}]{ousidhoum2021probing}
Ousidhoum, N.; Zhao, X.; Fang, T.; Song, Y.; and Yeung, D.-Y. 2021.
\newblock Probing toxic content in large pre-trained language models.
\newblock In \emph{Proceedings of the 59th Annual Meeting of the Association for Computational Linguistics and the 11th International Joint Conference on Natural Language Processing (Volume 1: Long Papers)}, 4262--4274.

\bibitem[{Panickssery et~al.(2023)Panickssery, Gabrieli, Schulz, Tong, Hubinger, and Turner}]{panickssery2023steering}
Panickssery, N.; Gabrieli, N.; Schulz, J.; Tong, M.; Hubinger, E.; and Turner, A.~M. 2023.
\newblock Steering llama 2 via contrastive activation addition.
\newblock \emph{arXiv preprint arXiv:2312.06681}.

\bibitem[{Park, Choe, and Veitch(2023)}]{park2023linear}
Park, K.; Choe, Y.~J.; and Veitch, V. 2023.
\newblock The linear representation hypothesis and the geometry of large language models.
\newblock \emph{arXiv preprint arXiv:2311.03658}.

\bibitem[{Paschalides, Pallis, and Dikaiakos(2025)}]{paschalides2025probing}
Paschalides, D.; Pallis, G.; and Dikaiakos, M.~D. 2025.
\newblock Probing the Subtle Ideological Manipulation of Large Language Models.
\newblock \emph{arXiv preprint arXiv:2504.14287}.

\bibitem[{Piao et~al.(2025)Piao, Lu, Gao, Xu, Santos, Li, and Evans}]{piao2025emergence}
Piao, J.; Lu, Z.; Gao, C.; Xu, F.; Santos, F.~P.; Li, Y.; and Evans, J. 2025.
\newblock Emergence of human-like polarization among large language model agents.
\newblock \emph{arXiv preprint arXiv:2501.05171}.

\bibitem[{Potter et~al.(2024)Potter, Lai, Kim, Evans, and Song}]{potter2024hidden}
Potter, Y.; Lai, S.; Kim, J.; Evans, J.; and Song, D. 2024.
\newblock Hidden Persuaders: LLMs' Political Leaning and Their Influence on Voters.
\newblock \emph{arXiv preprint arXiv:2410.24190}.

\bibitem[{Pres et~al.(2024)Pres, Ruis, Lubana, and Krueger}]{pres2024towards}
Pres, I.; Ruis, L.; Lubana, E.~S.; and Krueger, D. 2024.
\newblock Towards Reliable Evaluation of Behavior Steering Interventions in LLMs.
\newblock \emph{arXiv preprint arXiv:2410.17245}.

\bibitem[{R{\"o}ttger et~al.(2024)R{\"o}ttger, Hofmann, Pyatkin, Hinck, Kirk, Sch{\"u}tze, and Hovy}]{rottger2024political}
R{\"o}ttger, P.; Hofmann, V.; Pyatkin, V.; Hinck, M.; Kirk, H.~R.; Sch{\"u}tze, H.; and Hovy, D. 2024.
\newblock Political compass or spinning arrow? towards more meaningful evaluations for values and opinions in large language models.
\newblock \emph{arXiv preprint arXiv:2402.16786}.

\bibitem[{Rozado(2024)}]{rozado2024political}
Rozado, D. 2024.
\newblock The political preferences of LLMs.
\newblock \emph{PloS one}, 19(7): e0306621.

\bibitem[{Subramani, Suresh, and Peters(2022)}]{subramani2022extracting}
Subramani, N.; Suresh, N.; and Peters, M.~E. 2022.
\newblock Extracting latent steering vectors from pretrained language models.
\newblock \emph{arXiv preprint arXiv:2205.05124}.

\bibitem[{Tak et~al.(2025)Tak, Banayeeanzade, Bolourani, Kian, Jia, and Gratch}]{tak2025mechanistic}
Tak, A.~N.; Banayeeanzade, A.; Bolourani, A.; Kian, M.; Jia, R.; and Gratch, J. 2025.
\newblock Mechanistic Interpretability of Emotion Inference in Large Language Models.
\newblock \emph{arXiv preprint arXiv:2502.05489}.

\bibitem[{Tan and Lee(2025)}]{tan2025unmasking}
Tan, B. C.~Z.; and Lee, R. K.-W. 2025.
\newblock Unmasking Implicit Bias: Evaluating Persona-Prompted LLM Responses in Power-Disparate Social Scenarios.
\newblock \emph{arXiv preprint arXiv:2503.01532}.

\bibitem[{Team et~al.(2025)Team, Kamath, Ferret, Pathak, Vieillard, Merhej, Perrin, Matejovicova, Ram{\'e}, Rivi{\`e}re et~al.}]{team2025gemma}
Team, G.; Kamath, A.; Ferret, J.; Pathak, S.; Vieillard, N.; Merhej, R.; Perrin, S.; Matejovicova, T.; Ram{\'e}, A.; Rivi{\`e}re, M.; et~al. 2025.
\newblock Gemma 3 technical report.
\newblock \emph{arXiv preprint arXiv:2503.19786}.

\bibitem[{Turner et~al.(2023{\natexlab{a}})Turner, Thiergart, Leech, Udell, Vazquez, Mini, and MacDiarmid}]{turner2023activation}
Turner, A.~M.; Thiergart, L.; Leech, G.; Udell, D.; Vazquez, J.~J.; Mini, U.; and MacDiarmid, M. 2023{\natexlab{a}}.
\newblock Activation addition: Steering language models without optimization.
\newblock \emph{arXiv}, arXiv:2308.

\bibitem[{Turner et~al.(2023{\natexlab{b}})Turner, Thiergart, Leech, Udell, Vazquez, Mini, and MacDiarmid}]{turner2023steering}
Turner, A.~M.; Thiergart, L.; Leech, G.; Udell, D.; Vazquez, J.~J.; Mini, U.; and MacDiarmid, M. 2023{\natexlab{b}}.
\newblock Steering language models with activation engineering.
\newblock \emph{arXiv preprint arXiv:2308.10248}.

\bibitem[{Vogel(2024)}]{vogel2024repeng}
Vogel, T. 2024.
\newblock repeng.

\bibitem[{Wan et~al.(2023)Wan, Pu, Sun, Garimella, Chang, and Peng}]{wan-etal-2023-kelly}
Wan, Y.; Pu, G.; Sun, J.; Garimella, A.; Chang, K.-W.; and Peng, N. 2023.
\newblock {\textquotedblleft}Kelly is a Warm Person, Joseph is a Role Model{\textquotedblright}: Gender Biases in {LLM}-Generated Reference Letters.
\newblock In Bouamor, H.; Pino, J.; and Bali, K., eds., \emph{Findings of the Association for Computational Linguistics: EMNLP 2023}, 3730--3748. Singapore: Association for Computational Linguistics.

\bibitem[{Wojcik, Cislak, and Schmidt(2021)}]{wojcik2021left}
Wojcik, A.~D.; Cislak, A.; and Schmidt, P. 2021.
\newblock ‘The left is right’: Left and right political orientation across Eastern and Western Europe.
\newblock \emph{The Social Science Journal}, 1--17.

\bibitem[{Wu et~al.(2025)Wu, Arora, Geiger, Wang, Huang, Jurafsky, Manning, and Potts}]{wu2025axbenchsteeringllmssimple}
Wu, Z.; Arora, A.; Geiger, A.; Wang, Z.; Huang, J.; Jurafsky, D.; Manning, C.~D.; and Potts, C. 2025.
\newblock AxBench: Steering LLMs? Even Simple Baselines Outperform Sparse Autoencoders.
\newblock arXiv:2501.17148.

\bibitem[{Wu et~al.(2024)Wu, Geiger, Arora, Huang, Wang, Goodman, Manning, and Potts}]{wu2024pyvene}
Wu, Z.; Geiger, A.; Arora, A.; Huang, J.; Wang, Z.; Goodman, N.~D.; Manning, C.~D.; and Potts, C. 2024.
\newblock pyvene: A library for understanding and improving pytorch models via interventions.
\newblock \emph{arXiv preprint arXiv:2403.07809}.

\bibitem[{Xu et~al.(2024)Xu, Huang, Chen, and Wang}]{xu2024uncovering}
Xu, Z.; Huang, R.; Chen, C.; and Wang, X. 2024.
\newblock Uncovering safety risks of large language models through concept activation vector.
\newblock \emph{Advances in Neural Information Processing Systems}, 37: 116743--116782.

\bibitem[{Yang et~al.(2025)Yang, Li, Yang, Zhang, Hui, Zheng, Yu, Gao, Huang, Lv, Zheng, Liu, Zhou, Huang, Hu, Ge, Wei, Lin, Tang, Yang, Tu, Zhang, Yang, Yang, Zhou, Zhou, Lin, Dang, Bao, Yang, Yu, Deng, Li, Xue, Li, Zhang, Wang, Zhu, Men, Gao, Liu, Luo, Li, Tang, Yin, Ren, Wang, Zhang, Ren, Fan, Su, Zhang, Zhang, Wan, Liu, Wang, Cui, Zhang, Zhou, and Qiu}]{yang2025qwen3technicalreport}
Yang, A.; Li, A.; Yang, B.; Zhang, B.; Hui, B.; Zheng, B.; Yu, B.; Gao, C.; Huang, C.; Lv, C.; Zheng, C.; Liu, D.; Zhou, F.; Huang, F.; Hu, F.; Ge, H.; Wei, H.; Lin, H.; Tang, J.; Yang, J.; Tu, J.; Zhang, J.; Yang, J.; Yang, J.; Zhou, J.; Zhou, J.; Lin, J.; Dang, K.; Bao, K.; Yang, K.; Yu, L.; Deng, L.; Li, M.; Xue, M.; Li, M.; Zhang, P.; Wang, P.; Zhu, Q.; Men, R.; Gao, R.; Liu, S.; Luo, S.; Li, T.; Tang, T.; Yin, W.; Ren, X.; Wang, X.; Zhang, X.; Ren, X.; Fan, Y.; Su, Y.; Zhang, Y.; Zhang, Y.; Wan, Y.; Liu, Y.; Wang, Z.; Cui, Z.; Zhang, Z.; Zhou, Z.; and Qiu, Z. 2025.
\newblock Qwen3 Technical Report.
\newblock arXiv:2505.09388.

\bibitem[{Zhao et~al.(2024)Zhao, Zhao, Shen, Payani, Yang, and Du}]{zhao2024beyond}
Zhao, H.; Zhao, H.; Shen, B.; Payani, A.; Yang, F.; and Du, M. 2024.
\newblock Beyond single concept vector: Modeling concept subspace in llms with gaussian distribution.
\newblock \emph{arXiv preprint arXiv:2410.00153}.

\bibitem[{Zou et~al.(2023)Zou, Phan, Chen, Campbell, Guo, Ren, Pan, Yin, Mazeika, Dombrowski et~al.}]{zou2023representation}
Zou, A.; Phan, L.; Chen, S.; Campbell, J.; Guo, P.; Ren, R.; Pan, A.; Yin, X.; Mazeika, M.; Dombrowski, A.-K.; et~al. 2023.
\newblock Representation engineering: A top-down approach to ai transparency.
\newblock \emph{arXiv preprint arXiv:2310.01405}.

\end{thebibliography}

\clearpage
\appendix
\section{Appendix}

\subsubsection{Related Work}
Table \ref{tab:cmp_diff_poli} compares recent work with ours, and here are further clarifications about the listed statistics. 
We constructed the dataset ($\sim10$K) by combining 970 news events with opinion terms relevant to each event’s topic (see Table \ref{tab:concept_set}).
\citet{rottger2024political} is a prompt-based study, so it does not require additional training data. For Politune \cite{agiza2024politune}, we refer to the preference datasets they released on the Hugging Face Hub as their data size, including 2,356 left-leaning\footnote{\url{https://huggingface.co/datasets/scale-lab/politune-left}} and 2,825 right-leaning\footnote{\url{https://huggingface.co/datasets/scale-lab/politune-right}} samples, with a total of 5,182 ($\sim5,000$).
For \cite{kim2025linear}, we refer to the mentioned N = 552 prompts as the data size ($\sim500$). 


\subsubsection{Models and Data}
Table \ref{tab:lllm_sts} lists eight open-source models explored in this work. 

\begin{table}[http]
\centering
\caption{The Summary of LLMs Statistics}
\label{tab:lllm_sts}
\resizebox{0.48\textwidth}{!}{\begin{tabular}{cccc}
\hline
\textbf{Family} & \textbf{Model Name} & \textbf{Layers} & \textbf{Template} \\
\hline
Meta      & Llama3-1B   & 16 & \texttt{<|begin\_of\_text|>} \\
          & Llama3-3B   & 28 & \texttt{<|begin\_of\_text|>} \\
          & Llama3-8B   & 32 & \texttt{<|begin\_of\_text|>} \\
\hline
Google     & Gemma-1B    & 26 & \texttt{<start\_of\_turn>} \\
          & Gemma-7B    & 38 & \texttt{<start\_of\_turn>} \\
\hline
Qwen      & Qwen3-4B    & 36 & \texttt{<|im\_start|>} \\
          & Qwen3-8B    & 36 & \texttt{<|im\_start|>} \\
\hline
Mistralai & Mistral-7B  & 32 & \texttt{[INST]} \\
\hline
\end{tabular}}
\end{table}
Table \ref{tab:concept_set} lists the terms corresponding to left- and right-leaning across the different dimensions.
We combine these terms with the topics under each dimension to prompt GPT-4o to generate different learning statements on the given topics, thereby constructing the dataset. Table 6 shows an example of such statements. The final prompts are based on each model's corresponding template—for example, the template to the Mistral-7B model uses the pattern `[INST] [Constructed Statement] [/INST]'.

\subsubsection{Learning Task}
\begin{figure*}[t]
    \centering
    \includegraphics[width=0.73\linewidth]{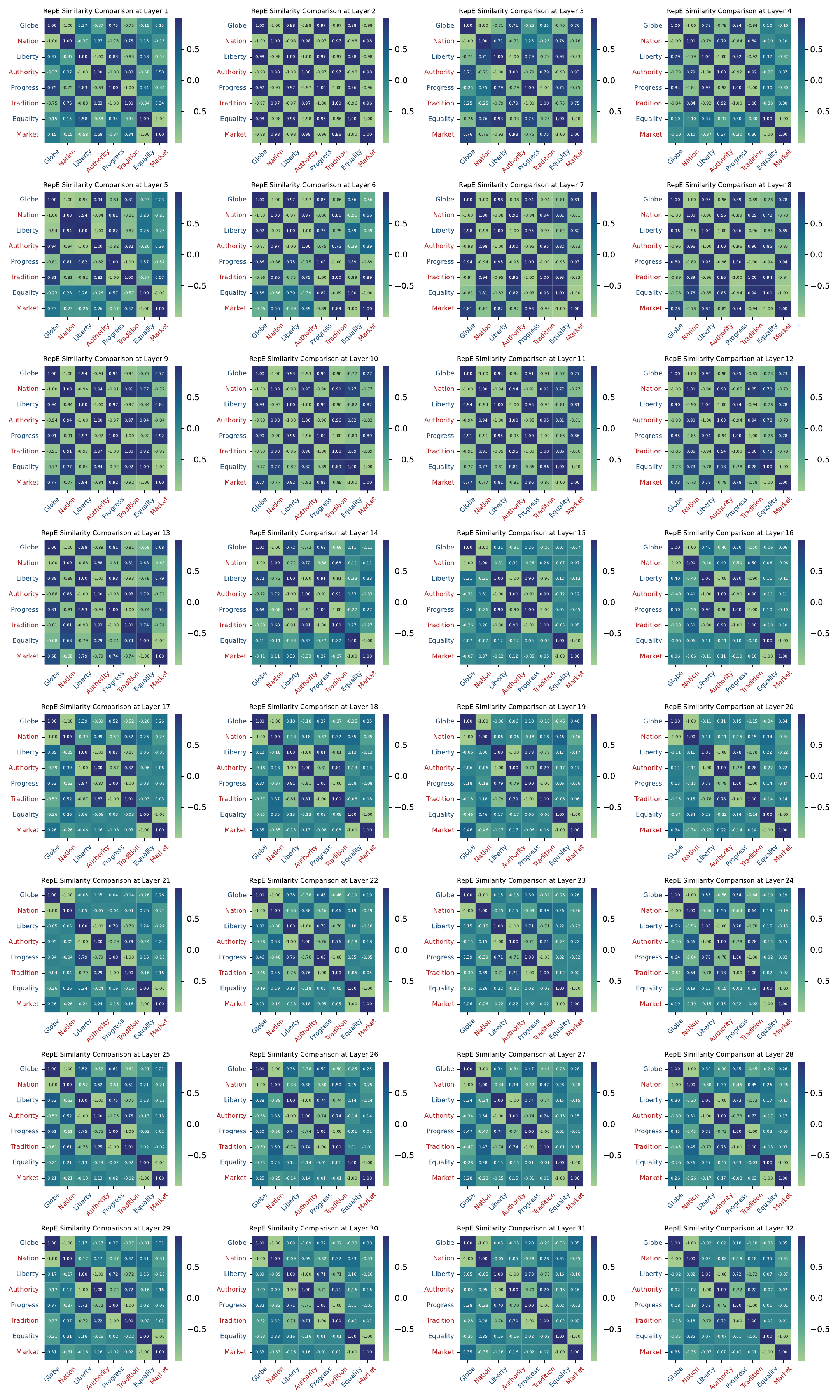}
    \caption{RepE Political Concept Vectors Correlations across Layers (Llama3-8B)}
    \label{fig:repe_corr_disentangle}
\end{figure*}

Figure \ref{fig:repe_corr_disentangle} is a correlation heatmap of our learned political concept vectors across layers. We follow the procedure in \cite{chen2024designing} for vector learning. The difference between the vectors used for detection and intervention tasks is that, for detection, we append the cue phrase `The leaning is' to trigger political-concept detection, whereas this is omitted in the intervention task.

\subsubsection{Detection Task}
Table \ref{tab:best_cmp} shows the best detection performance of three methods (CAA, RepE, Linear Probing) on dataset across different dimensions.

\subsubsection{Intervention Task}

\begin{figure*}[t]
    \centering    \includegraphics[width=1\linewidth]{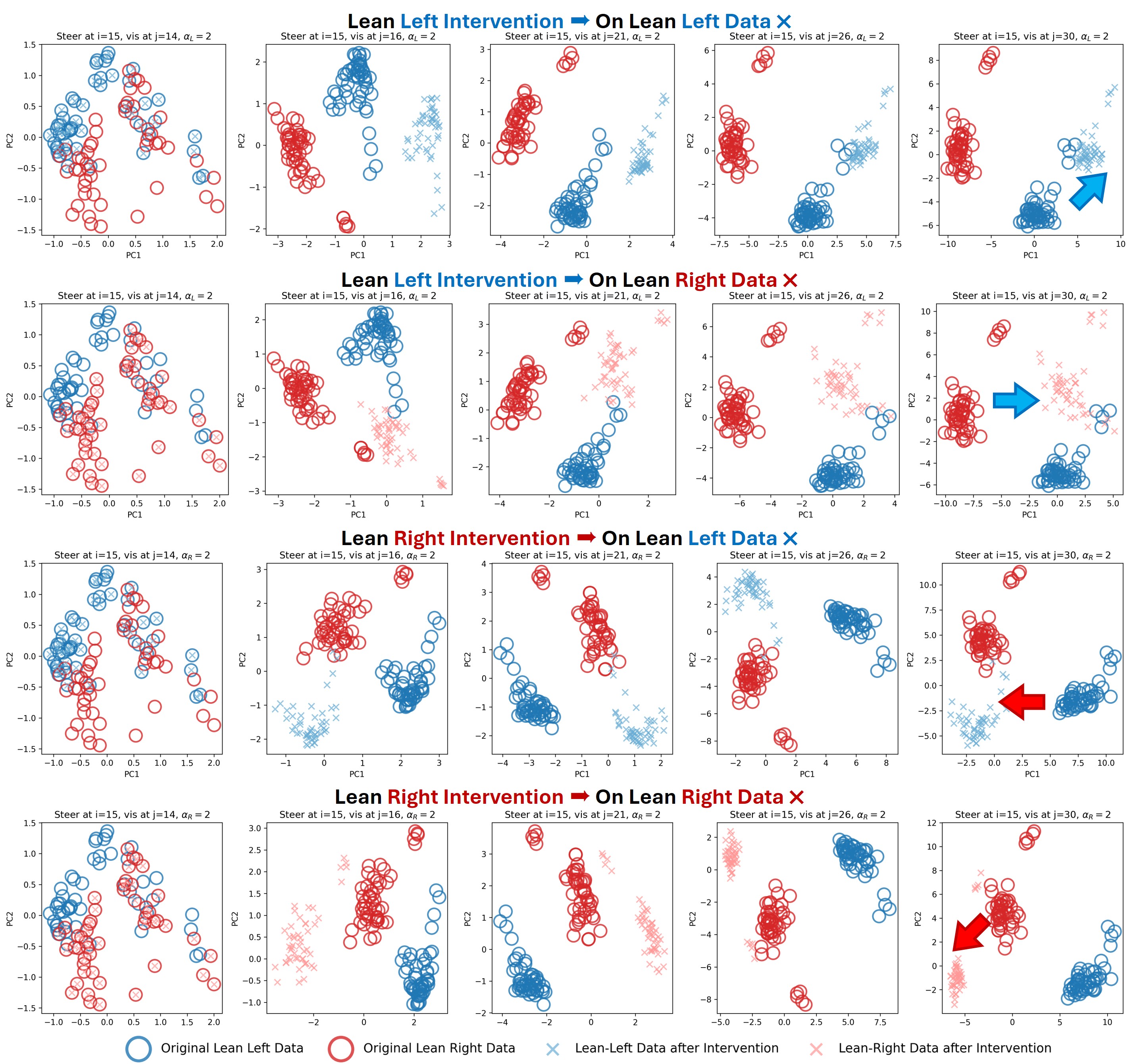}
    \caption{Single Layer Intervention. The circles $\circ$ refer to the original data distribution, while the crosses $\times$ show the distribution after the intervention on the original data. The colors denote political leaning (\textcolor{NavyBlue}{left}/\textcolor{BrickRed}{right}). When the intervention direction opposes the data's political leaning, the data embedding shifts toward the intervention direction. Specifically, applying a right-leaning intervention to left-leaning data causes the data to shift toward the right-leaning distribution area, while applying a left-leaning intervention to right-leaning data causes it to shift toward the left-leaning area. When the intervention and data share the same political leaning, the data distribution shifts but remains within the corresponding political leaning area.}
    \label{fig:single_inter_all}
\end{figure*}

To improve the readability of the images, the visualization of the distribution shift after single-layer intervention in Figure \ref{fig:single_interven} uses the first 50 left statements and the first 50 right statements in the test set. Figure \ref{fig:single_inter_all} shows the detailed single-layer interventions at layer $\ell=15$ with strength ($\alpha_{L}=2, \alpha_{R}=2$). In response intervention, the LLMs penalty parameter was set in the range of 1 to 1.4, with the temperature fixed at 0.2, and the maximum number of response tokens set to 100.

The multi-layer intervention experiments show that our learned concept vectors can steer LLM outputs.
Nonetheless, there exist certain cases (e.g., healthcare discussions) where LLMs resist steering. One possible cause relates to the data source: although we collected many events (news) on diverse topics from AllSides, they are mainly U.S. news. As a result, the LLMs' representations may embed U.S.-specific concepts, pulling the learned vectors toward more American cultures.
An example is that healthcare systems differ worldwide. Some European countries provide universal healthcare, whereas the U.S. relies mainly on private healthcare—economically, the former is more left-leaning (equality), while the latter is more right-leaning (market). therefore, if U.S. economic news dominates the corpus, market-oriented concept may influence certain topics (e.g., healthcare).
Within our current fine-grained framework, incorporating national factors into consideration is a worthwhile direction for future work.

\begin{table}[http]
\caption{The Best Detection Performance of CAA, RepE, Linear Probing on Different Dimensions, where Prob Refers to Linear Probing.}
\label{tab:best_cmp}
\begin{tabular}{ccccc}
\hline
\textbf{Model Names }                                & \textbf{Dimensions} & \textbf{CAA} & \textbf{RepE} & \textbf{Prob} \\
\hline
\multirow{4}{*}{\textbf{Llama3-1B}}  & Economic   & 0.8066 & 0.5610 & 0.9574 \\
                                     & Diplomatic & 0.8340 & 0.5233 & 0.9645 \\
                                     & Civil      & 0.6404 & 0.5130 & 0.8677 \\
                                     & Society    & 0.7052 & 0.5219 & 0.9234 \\ \hline
\multirow{4}{*}{\textbf{Llama3-3B}}  & Economic   & 0.7463 & 0.5647 & 0.9691 \\
                                     & Diplomatic & 0.8456 & 0.5097 & 0.9793 \\
                                     & Civil      & 0.6396 & 0.5317 & 0.8929 \\
                                     & Society    & 0.7385 & 0.5188 & 0.9396 \\ \hline
\multirow{4}{*}{\textbf{Llama3-8B}}  & Economic   & 0.9706 & 0.9412 & 0.9868 \\
                                     & Diplomatic & 0.9819 & 0.8301 & 0.9910 \\
                                     & Civil      & 0.8677 & 0.7946 & 0.9156 \\
                                     & Society    & 0.8714 & 0.8078 & 0.9651 \\ \hline
\multirow{4}{*}{\textbf{Gemma-1B}}   & Economic   & 0.8993 & 0.5015 & 0.9463 \\
                                     & Diplomatic & 0.5097 & 0.5090 & 0.9632 \\
                                     & Civil      & 0.5455 & 0.5000 & 0.8506 \\
                                     & Society    & 0.6203 & 0.5010 & 0.9104 \\ \hline
\multirow{4}{*}{\textbf{Gemma-7B}}   & Economic   & 0.9074 & 0.5647 & 0.9676 \\
                                     & Diplomatic & 0.9490 & 0.9444 & 0.9767 \\
                                     & Civil      & 0.7857 & 0.5706 & 0.9067 \\
                                     & Society    & 0.8474 & 0.5589 & 0.9453 \\ \hline
\multirow{4}{*}{\textbf{Qwen3-4B}}   & Economic   & 0.8809 & 0.6169 & 0.9625 \\
                                     & Diplomatic & 0.8766 & 0.6686 & 0.9787 \\
                                     & Civil      & 0.7159 & 0.5966 & 0.9010 \\
                                     & Society    & 0.7578 & 0.5682 & 0.9464 \\ \hline
\multirow{4}{*}{\textbf{Qwen3-8B}}   & Economic   & 0.9515 & 0.9353 & 0.9787 \\
                                     & Diplomatic & 0.9748 & 0.8559 & 0.9858 \\
                                     & Civil      & 0.8369 & 0.7573 & 0.9229 \\
                                     & Society    & 0.8984 & 0.8042 & 0.9542 \\ \hline
\multirow{4}{*}{\textbf{Mistral-7B}} & Economic   & 0.9346 & 0.5831 & 0.9721 \\
                                     & Diplomatic & 0.9477 & 0.5472 & 0.9774 \\
                                     & Civil      & 0.8101 & 0.5284 & 0.8994 \\
                                     & Society    & 0.8859 & 0.5167 & 0.9521 \\
 \hline
\end{tabular}
\end{table}

\begin{table*}[t]
\caption{The Details of Collected Concept Set}
\label{tab:concept_set}
\resizebox{0.95\textwidth}{!}{
\begin{tabular}{ccccc}
\hline
\textbf{Dimension}                    & \textbf{Leaning} & \textbf{Concept}   & \textbf{Sub-Level Concept Examples}                                                & \textbf{Total} \\
\hline
\multirow{2}{*}{Economic} & Left & Equality & Progressive tax, regualtion, planned economy,   wealth redistribution & 10 \\
                            & Right    & Market    & Flat tax, deregulation,  laissez-faire, wealth accumulation & 10    \\
\hline
\multirow{2}{*}{Diplomatic} & Left    & Globe     & World government, immigration,   diplomacy, globalist                       & 12    \\
                            & Right    & Nation    & Sovereignty, border control, military   strength, notionalist               & 12    \\
\hline
\multirow{2}{*}{Civil}    & Left & Liberty  & Checks and balances, autonomy, free   expression        & 16 \\
                            & Right    & Authority & Centralization, regulation,   censorship                      & 16    \\
\hline
\multirow{2}{*}{Society}    & Left    & Progress  & Reason, development, scientific,   technology                               & 12    \\
                            & Right    & Tradition & Moral, status quo, religion, natural                                        & 12   \\
\hline
\end{tabular}}
\end{table*}

\begin{table*}[t]
    \centering
    \caption{Some Examples of the Generated Data}
    \includegraphics[width=1\linewidth]{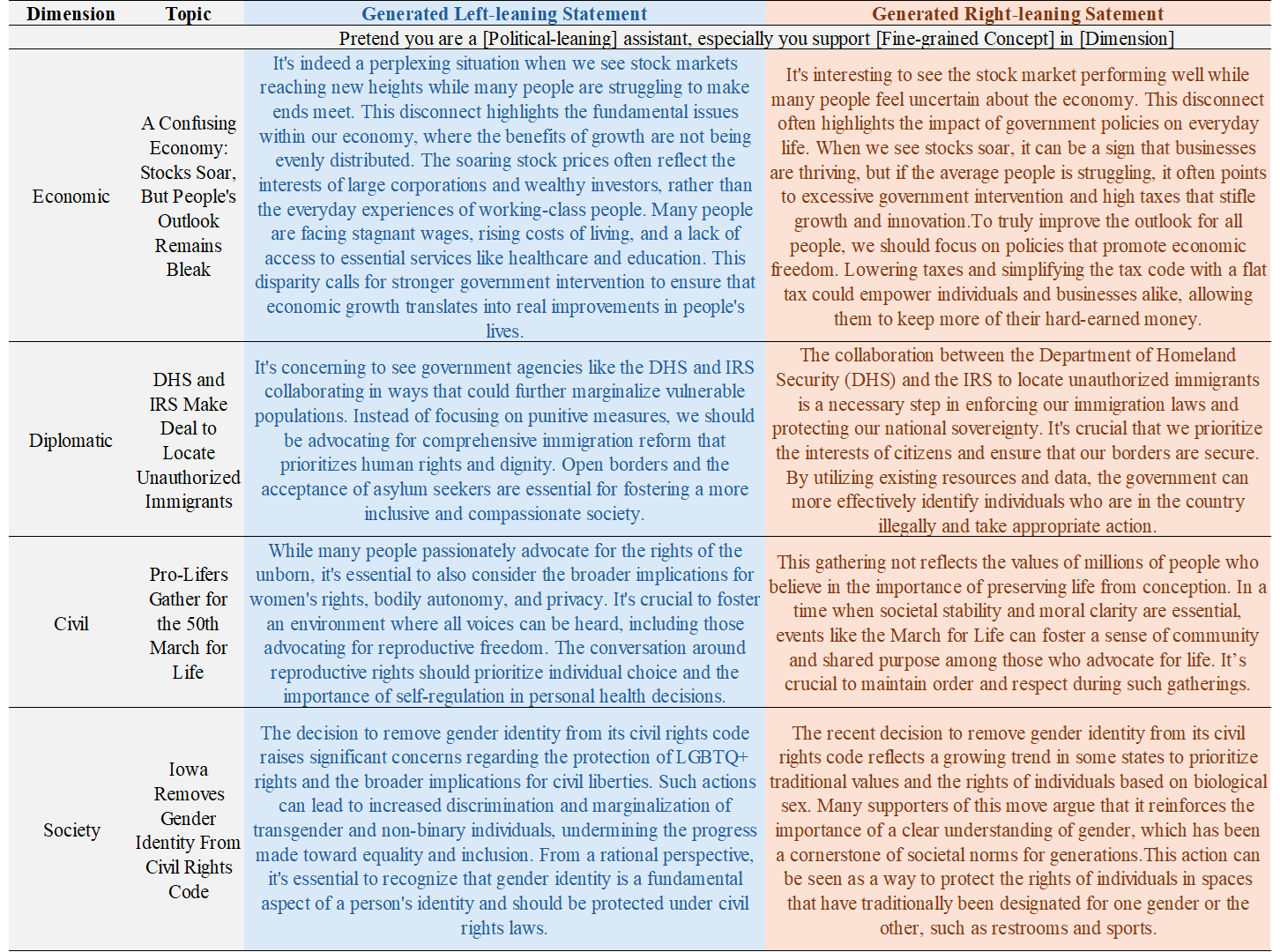}
\end{table*}

\end{document}